\documentclass{article}
\usepackage{arxiv}
\usepackage{times}  % DO NOT CHANGE THIS
\usepackage{helvet}  % DO NOT CHANGE THIS
\usepackage{courier}  % DO NOT CHANGE THIS
\usepackage[hyphens]{url}  % DO NOT CHANGE THIS
\usepackage{graphicx} % DO NOT CHANGE THIS
\usepackage[square,numbers]{natbib}  % DO NOT CHANGE THIS AND DO NOT ADD ANY OPTIONS TO IT
\usepackage{caption} % DO NOT CHANGE THIS AND DO NOT ADD ANY OPTIONS TO IT
\usepackage{algorithm}

\usepackage{doi}

\usepackage{newfloat}
\usepackage{listings}
\DeclareCaptionStyle{ruled}{labelfont=normalfont,labelsep=colon,strut=off} % DO NOT CHANGE THIS
\floatstyle{ruled}
\newfloat{listing}{tb}{lst}{}
\floatname{listing}{Listing}
\title{A Unified Framework for Neural Computation and Learning Over Time}

\author{Stefano Melacci \\
	DIISM\\
	University of Siena, Italy \\
	\texttt{\footnotesize stefano.melacci@unisi.it} \\
	\And
	Alessandro Betti \\
	Scuola IMT Alti Studi Lucca\\
	Lucca, Italy \\
	\texttt{\footnotesize alessandro.betti@imtlucca.it} \\
        \And
	Michele Casoni \\
	DIISM\\
	University of Siena, Italy \\
	\texttt{\footnotesize m.casoni@student.unisi.it} \\
        \And
	Tommaso Guidi \\
	DINFO\\
	University of Florence, Italy \\
	\texttt{\footnotesize tommaso.guidi@unifi.it} \\ 
        \And        
	Matteo Tiezzi \\
	Italian Institute of Technology\\
	Genoa, Italy \\
	\texttt{\footnotesize matteo.tiezzi@iit.it} \\ 
        \And        
	Marco Gori \\
	DIISM\\
	University of Siena, Italy \\
	\texttt{\footnotesize marco.gori@unisi.it} \\
}

\usepackage{bibentry}
\usepackage{amsfonts} 
\usepackage{amsmath}
\usepackage{algpseudocode}
\usepackage{caption}
\usepackage{pdfrender}
\usepackage{color}  % AAAI DOES NOT WANT THIS!!!!

\newcommand\parafango[1]{\vskip 0.5mm \noindent\textbf{#1.
\enspace}}

\newcommand*{\boldgreek}[1]{%
  \textpdfrender{%
    TextRenderingMode=FillStroke,%
    LineWidth=.4pt,%
  }{#1}%
}

\def\w{{\boldgreek{\theta}}}
\def\W{{\boldgreek{\Theta}}}
\def\y{\mathrm{\bf y}}
\renewcommand{\xi}{{\mathrm{\bf h}}}  % do not touch this!
\let\phiphi\phi  % change this one if you want to change the meaning of \phi
\renewcommand{\phi}{\phiphi}  % do not touch this!
\def\phiexp{\phi}
\def\eulerstep{\tau}
\def\t{t}
\def\k{\kappa}
\def\up{\upsilon}
\def\s{s}
\def\tprev{\t-\eulerstep}
\def\tnext{\t+\eulerstep}

\def\argt{_{\t}}
\def\argti{_{i,\t}}

\def\argk{_{\k}}
\def\argu{_{\up}}

\def\knext{\k+1}
\def\kprev{\k-1}
\def\unext{\up+1}

\def\argtnext{_{\tnext}}

\def\argtprev{_{\tprev}}

\def\argknext{_{\knext}}
\def\argkprev{_{\kprev}}
\def\argunext{_{\unext}}

\def\argzero{(0)}
\def\tA{0}
\def\tB{N}
\def\flipper{s}
\def\dis{\eta}
\def\numlayers{\nu}
\def\costate{{\bf p}}
\DeclareMathOperator*{\argmin}{arg\,min}

\def\argpsit{_{\psi\argt}}
\def\argpsitnext{_{\psi\argtnext}}
\def\argpsitprev{_{\psi\argtprev}}

\def\x{\mathrm{\bf x}}
\def\xxi{\xi}
\def\xw{\w}
\def\xwxi{\w^{\xi}}
\def\xwy{\w^{\y}}
\def\dotx{\dot{\x}}
\def\dotxxi{\dot{\xxi}}
\def\dotxwxi{\dot{\w}^{\f}}
\newcommand{\fastxxib}[1]{\underline{\x}^{\xi,(#1)}}

\newcommand{\yb}[1]{\y^{(#1)}}

\newcommand{\xxib}[1]{\xi^{(#1)}}

\newcommand{\dotxxib}[1]{\dot{\xi}^{(#1)}}

\usepackage{accents}

\def\p{\hat{\x}}
\def\pxi{\mathcal{\bf z}}
\def\pw{\boldgreek{\omega}}
\def\pwxi{\boldgreek{\omega}^{\h}}
\def\pwy{\boldgreek{\omega}^{\y}}
\def\dotxw{\dot{\w}}
\def\dotp{\dot{\p}}
\def\dotpxi{\dot{\pxi}}
\def\dotpw{\dot{\pw}}
\def\dotpwxi{\dot{\pw}^{\h}}
\def\dotpwy{\dot{\pw}^{\y}}

\def\controlset{A}

\newcommand{\pb}[1]{p^{(#1)}}
\newcommand{\pxib}[1]{\pxi^{(#1)}}

\newcommand{\dotpxib}[1]{\dot{\p}^{\xi,(#1)}}

\def\h{\mathrm{\bf h}}

\def\f{f}
\def\act{\sigma}

\def\fxi{{\f}^{\xi}}

\def\fw{\bar{\w}}

\def\fwtmp{\underline{\w}}

\def\fxid{{\hat{\f}^{\xi}}}

\def\fy{\f^{\y}}

\def\H{H}
\def\Hrest{\H^{\prime}}

\def\inneurons{\mathcal{I}}
\def\outneurons{\mathcal{O}}
\def\hiddenneurons{\mathcal{H}}
\def\plainneurons{\mathcal{P}}
\def\recurrentneurons{\mathcal{R}}

\def\agent{\mathfrak{A}}
\def\stream{\mathcal{S}}
\def\u{\mathrm{\bf u}}
\def\target{\hat{\mathrm{\bf y}}}

\newcommand{\ub}[1]{\u^{(#1)}}
\newcommand{\targetb}[1]{\hat{\y}^{(#1)}}

\def\inspace{\mathbb{I}}
\def\outspace{\mathbb{O}}
\def\weightspace{\mathbb{W}}
\def\targetspace{\mathbb{T}}
\def\realset{\mathbb{R}}
\def\indicator{\mathrm{\mathbf{1}}}
\def\negindicator{\mathrm{\mathbf{0}}}

\def\betaw{{\boldgreek{\beta}}}
\def\betawxi{\betaw^{\xi}}
\def\betawy{\betaw^{\y}}
\def\lr{\gamma}
\def\lrwy{\lr^{\w,\y}}
\def\g{\mathrm{\bf g}}
\renewcommand\paragraph[1]{\noindent\textbf{#1
\enspace}}
\begin{document}

\maketitle

\begin{abstract}
This paper proposes Hamiltonian Learning, a novel unified framework for learning with neural networks ``over time'', i.e., from a possibly infinite stream of data, in an online manner, without having access to future information.
Existing works focus on the simplified setting in which the stream has a known finite length or is segmented into smaller sequences, leveraging well-established learning strategies from statistical machine learning.  In this paper, the problem of learning over time is rethought from scratch, leveraging tools from optimal control theory, which yield a unifying view of the temporal dynamics of neural computations and learning. Hamiltonian Learning is based on differential equations that: ($i$) can be integrated without the need of external software solvers; ($ii$) generalize the well-established notion of gradient-based learning in feed-forward and recurrent networks; ($iii$) open to novel perspectives. The proposed framework is showcased by experimentally proving how it can recover gradient-based learning, comparing it to out-of-the box optimizers, and describing how it is flexible enough to switch from fully-local to partially/non-local computational schemes, possibly distributed over multiple devices, and BackPropagation without storing activations. Hamiltonian Learning is easy to implement and can help researches approach in a principled and innovative manner the problem of learning over time.
\end{abstract}

\section{Introduction}
\label{sec:introduction}
\parafango{Motivations} A longstanding challenge in machine learning with neural networks is the one of designing models and learning strategies that are naturally conceived to learn ``over time'', progressively adapting to the information from a stream of data \cite{wang2024comprehensive,gori2023collectionless}. This implies dealing with possibly infinite streams, online learning, no access to future information, thus going beyond classic statistical methods exploiting offline-collected datasets.
%the importance of this learning setting might to go beyond privacy and geopolitical issues related to classic statistical methods that store large dataset and learn in an offline manner \cite{gori2023collectionless}. 
In this paper, tools from optimal control theory \cite{bertsekas2012dynamic} are exploited to rethink learning over time {\it from scratch}, proposing a unifying framework named Hamiltonian Learning (HL). 
Differential equations drive the learning dynamics, which are integrated going ``forward'' in time, i.e., without back-propagating to the past. Although related tools are widely used (in a different way) in reinforcement learning  \cite{khetarpal2022towards}, and there exist works exploiting control theory in the context of neural networks \cite{neuralode,neuralcde,araujo2023a}, to the best of our knowledge they have not addressed the general problem of learning over time from a continuous, possibly infinite, stream of data. A few recent exceptions do exist, but they are focused on specific cases with significant limitations \cite{betti2022continual,betti2023neural}. HL is rooted on a state-space formulation of the neural model, that yields computations which are fully local in space and time. The popularity of state-space models has become prominent in recent literature \cite{tiezzi2024resurgence,de2024griffin,gu2023mamba}, further motivating the HL perspective to create a strong connection between control theory and machine learning \cite{alonso2024state}.

\parafango{Contributions} The main contributions of this paper are the following. 
    ($i$) We propose a unified framework for neural computation and learning over time, exploiting well-established tools from optimal control theory. HL is designed to leverage the Hamilton Equations \cite{bertsekas2012dynamic} to learn in a forward manner, facing an initial-value problem, instead of a boundary-value problem (Section~\ref{sec:bg}--preliminaries, Section~\ref{sec:roastedham}--HL).
    ($ii$) We (formally and experimentally) show that, by integrating the differential equations with the Euler method, thus without any off-the-shelf solvers for differential equations, and enforcing a sequential constraint to the update operations (which are otherwise fully parallel), our framework recovers the most popular  gradient-based learning, i.e., BackPropagation and BackPropagation Through Time (Section~\ref{sec:grad}). %We showcase how this also holds when directly dealing with a discrete formulation of the problem, as it is common in applications.
    ($ii$) We discuss how HL provides a uniform and flexible view of neural computation over a stream of data, which is fully local in time and space. This favours customizability in terms of parallelization, distributed computation, and it also generalizes approaches to memory efficient BackPropagation (also Through Time) without storing outputs/activations (Section~\ref{sec:good}).
%%   
%\parafango{Perspectives} 
The generality of HL is not intended to provide tools to solve classic issues in lifelong/continual learning (e.g., catastrophic forgetting), but to provide researches with a flexible framework that, due to its novelty, generality, and accessibility, might open to novel achievement in learning over time, which is not as mature as offline learning. %HL does not require off-the-shelf solvers for differential equations.

\section{Preliminaries}
\label{sec:bg}
%This section introduces our learning setting and revisits popular concepts about neural architectures using an ad-hoc notation that will help present the ideas of this paper. 

%While our goal is to promote a compact view that covers the propagation of information through the network structure (i.e., through layers) and through time in a joint manner, we will initially keep these two aspects distinct, to merge them in Section~\ref{sec:good}.

\parafango{Notation} %We consider the time horizon $[\tA,\tB)$, where $\tB$ could be possibly infinite, being $t$ a generic time. Data $\u(t) \in \realset^{d}$ is provided to an agent implementing a neural network, %at (possibly unevenly spaced) instants $\{t_0, t_1, \ldots, \infty \}$, 
At time $\t \geq 0$ data $\u\argt \in \realset^{d}$ is provided to a neural network (for inference and learning), with learnable parameters collected into vector $\w$, with $\t \in [0, N)$, $N >0$, that could be possibly $\infty$. 
Our framework is devised in a continuous-time setting, which, as usual, is implemented by evaluating functions at discrete time instants, that might be not evenly spaced. We avoid introducing further notation to formalize the transition from continuous-time differential equations to the outcome of the discrete-time integration steps, which will be clear from the context. %Hence, we introduce the time variable $\t$ and use the compact notation $\u\argt$ to indicate $\u$ evaluated at time $\t$ (i.e., $\u\argk$ is the $\k$-th sample of the sequence, while $\u\argt$ is the sample at wall-clock time $\t$). %, where $\w_i\argt$ is the $i$-th element at time $\t$. 
%When dealing with discrete formulations, we will replace argument $(t)$ with subscript $\k \in \{0, 1, \ldots, \infty \}$ to refer to the step index, and $\w_{i,\k}$ is the $i$-th element of vector $\w$ at step $\k$. 
%We will introduce the time sometimes drop the time/step index to simplify the notation. 
%Neurons (including the $q$ input ones) are associated to unique integer identifiers in $\{0,\ldots,m-1\}$. %We will also use a matrix-based notation $\W \in \realset^{(m-q)\times m}$ to represent weights, where $\W_{ij}$ is the weight of the directed connection from the $j$-th neuron to the $i$-th one (it can be straightforwardly extended to include biases). The notation presented so far applies to all the vectors (they are column vectors--lowercase) and matrices (uppercase) of the paper, while scalar values and vector-valued functions are indicated with normal text, using uppercase letters to represent functions returning scalar values. 
% Calligraphy letters, e.g., $\hiddenneurons$, indicate sets.
Vectors are indicated in bold (they are column vectors--lowercase), where $\mathrm{\bf o}^{T}$ is the transpose of $\mathrm{\bf o}$.
The notation $[\mathrm{\bf o}, \u]$ is the concatenation of the comma-separated vectors in brackets. Given an $m$-real-valued function, e.g., $r(\u)$ returning vector $\mathrm{\bf o}$,
the Jacobian matrix with respect to its argument of length $d$ is a matrix in $\realset^{m\times d}$. As a consequence, gradients of scalar functions ($m=1$) are row vectors. The only exception to this rule is in the case of derivatives with respect to time, indicated with $\dot{\mathrm{\bf o}}$, that are still column vectors, as $\mathrm{\bf o}$. We will frequently follow the convention of  writing $\mathrm{\bf o}$ in place of $r(\u, \xi)$, keeping track of the dependencies with respect to the arguments of $r$. Dependencies do not propagate through time. 
%of an $m$-valued function $f$, $\mathrm{\bf o} = f(\u, \h, \w)$, with respect to its parameter $\x$ of length $n$ (for example), evaluated in $(\x^{\prime}, \w^{\prime}, \t^{\prime})$ is a matrix in $\realset^{m\times n}$ is either indicated with ${\partial f(\x^{\prime}, \w^{\prime}, \t^{\prime})}/{\partial\x^{\prime}}$ or with the shorthand notation ${\partial \mathrm{\bf o}}/{\partial\x^{\prime}}$, to keep it simple.
%As a consequence, gradients of scalar functions ($m=1$) are row vectors. The only exception to this rule is in the case of the derivative with respect to time, indicated with $\dot{\mathrm{\bf o}}$, that is still a column vector as $\mathrm{\bf o}$. 
%When considering batched computation, the superscript (in brackets) is the index of a batch element, such as $\ub{i}$. 
The operator $\odot$ is the Hadamard product.

\parafango{Feed-Forward and Recurrent Networks} A feed-forward neural model (including convolutional nets) is a (possibly deep) architecture that, given input $\u$, computes the values of the output neurons as $\h = {\f}(\u, \w)$.
%\begin{equation}
%    \begin{split}
%    \h &= {\f}(\u, \w),
%    \end{split}
%    \label{eq:f2}
%\end{equation}
%where $f$ can be iteratively implemented, computing the output of each neuron/layer accordingly to their topological ordering in the Directed Acyclic Graph (DAG) of the selected architecture. Vector $\bar\h$ collects some initial outputs of the neurons, which does not affect the computations (it can be anything). This way of formalizing inference in feed-forward networks, clearly redundant at this stage, will help describe the ideas in this paper. 
%Of course, as it is largely known, in multi-layer networks with $\numlayers$ layers it is enough to store the $\numlayers$ non-sparse blocks of $\W$ associated to the $\numlayers$ layers (i.e., the $\numlayers$ weight matrices), iterating over layers, and avoiding to repeat the whole $\W$-by-vector product at each fixed-point iteration. Residual units {\color{red}(citation)} can be easily modeled by adding DAG nodes in correspondence of the usual neuron-input addition operation which is typical residual connections. However, we can make them explicit by replacing $\sigma(\W [\u, \h])$ in Eq.~\ref{eq:f2} with $\Gamma\h + \sigma(\W [\u, \h])$, where $\Gamma$ is a binary matrix of the same size/organization of $\W$, with ones associated to residual connections.
Recurrent models are designed to process finite sequences of samples, such as $\u_0$, $\u_1$, $\ldots$, $\u_{n-1}$, % with one or more stacked recurrent layers or interleaved to dense layers ($\numlayers$ layers, overall). 
where we can evaluate the output values of the recurrent neurons at step $\knext$ as,
\begin{equation}
    \begin{split}
    \h\argknext &= {\f}(\u\argk, \h\argk, \w), \quad\text{for } \k=0,1,\ldots,n-1.
    \end{split}
    \label{eq:r2}
\end{equation}
In this case, computations depend on the initial $\h_0$ which is commonly set to zeros or to random values \cite{medsker2001recurrent}. Notice that $\w$ is not a function of $\k$, since the same weights are used to process the whole sequence. Moreover, Eq.~\ref{eq:r2} is a generalization of feed-forward networks, which are obtained by keeping $\k$ fixed and setting the second argument of $\f$ to $\mathrm{\bf 0}$ (vector of zeros). %For this reason, we will only refer to Eq.~\ref{eq:r2} in the following. 
Learning consists in determining the optimal value of $\w$, in the sense of minimizing a given loss function. For example, in the case of classic online gradient, weights are updated after having processed each example (feed-forward nets), or each sequence (recurrent nets). Formally, at a generic step $\upsilon$,
\begin{equation}
    \w\argunext = \w\argu - \lr \tilde{\w}\argu
    \label{eq:wwww}
\end{equation}
where $\lr > 0$ is the learning rate and $-\lr \tilde{\w}\argu$ is the variation.

\parafango{State and Stream} In this paper we consider an extended definition of state with respect to the one used in recurrent networks, $\h$. In particular, the state here is $[\h, \theta]$, which represents a snapshot of the model at a certain time instant, including all the information that is needed to compute Eq.~\ref{eq:r2} when an input is given.
%\parafango{Stream} %The perspective of this paper is the one in which learning consists of the online processing of a single stream of data, that could be possibly infinite. 
Differently from dataset-oriented approaches, here we consider a unique source of information, the stream $\stream$, that, at time $\t$, yields an input-target pair, $\stream\argt = (\u\argt, \target\argt)$,\footnote{Targets could also be present only for some $\t$'s.}.
%A given dataset of samples can be streamed one sample after the other, possibly randomizing the order in case of stochastic learning, and $\delta\argt=1$, $\forall \t$. If samples are sequences, then they are streamed one after the other. However, in this case each single sequence must be further streamed token-by-token, thus $\stream$ provides a triple with $\delta\argt = 1$ for the last token of the current sequence (otherwise $\delta\argt=0$), to preserve the information about the boundary between consecutively streamed sequences.
%The source of information is a stream $\stream$ that, at time $\t$, yields an input-target-tag triple, $\stream\argt = (\u\argt, \target\argt, \delta\argt)$,\footnote{Targets could also be present only for some $\t$'s.} where the binary tag $\delta$ is meaningful only if the stream is segmented, and it is $1$ for the last batch of a segment. 
%For example, a dataset of sequences becomes a unique long sequence composed of the concatenation of the contents of the dataset, which is progressively streamed. Each original sequence is a segment of $\stream$. 
We will avoid explicitly mentioning batched data, even if what we present in this paper is fine also in the case in which $\stream$ returns mini-batches.
%
%\parafango{Continuous Time} We will present our main results in a continuous time setting (time $\t$), that allows us to explicitly take care of the distance (in time) between consecutive samples from the stream, which might be uneven. However, we will also discuss discrete time, comparing the outcomes of the two settings. 
In practice, $\stream$ yields data only at specific time instants $\t_1, \t_2, \ldots, \t_{\infty}$, which may be evenly or unevenly spaced. To keep the notation simple, we will assume a consistent spacing of $\eulerstep$ between all consecutive samples, although our proposed method does not require evenly spaced samples (see also Appendix~\ref{further}).
%We assume the most basic schedule of computations: the agent which implements our neural networks monitors the stream and, when a batch of data is provided, it starts processing it, entering a ``busy'' state, and leaving it when it is done. Data from $\stream$ is discarded when the agent is busy. At time $\t_j$, the agent is aware of the time $\Delta(t_j)$ that passed from the previously processed sample.

\begin{figure}
    %\vskip -1mm
    %\vspace{0.5mm} 
    \centering
    \includegraphics[width=0.45\linewidth]{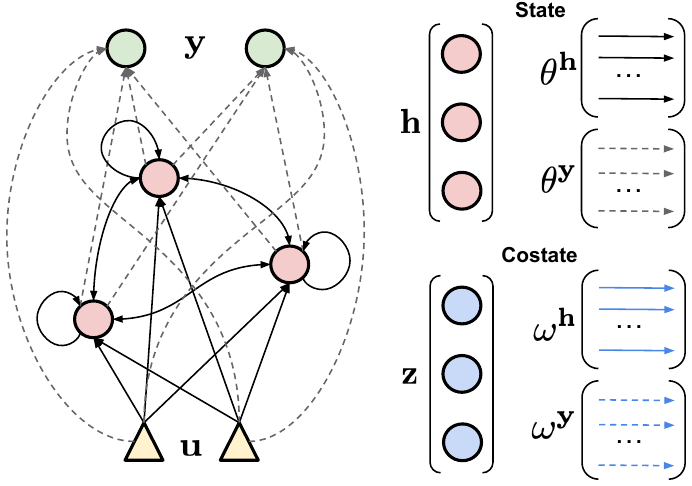}
    %\vskip -1.9mm
    \captionof{figure}{Left: State net (solid lines) and output net (dashed lines); Right: state ($\h$ and $\w=[\xwxi, \xwy]$). Costate ($\pxi$ and $\pw=[\pwxi,\pwy]$) is also shown (not part of the net).}
    \label{fig:toy}  
    %\vskip -3mm
\end{figure}

%\colorbox[RGB]{247,247,247}{
\begin{algorithm}
    %\footnotesize
    %\vspace{-0mm} 
    \begin{algorithmic}
    %\State \hrulefill \vspace{-0.5mm} 
    \State Init $\h_0$, $\w_0$, $\pxi_0$, $\pw_0$, $\t$. Select $\dis \geq 0$, $\betaw > 0$, and function $\phiexp\argt$.
    \State $\ $
    \While{$true$}
    \State $\ $ \vspace{-2mm}
    \State $(\u\argt, \target\argt) \gets \stream\argt$
    \State $\ $ \vspace{-2mm}
    \State $\dot{\mathrm{\bf h}}\argt \hskip-0.5mm=\hskip-0.5mm f^{\mathrm{\bf h}}(\mathrm{\bf u}\argt, \mathrm{\bf h}\argt, \mathrm{\bf \theta}\argt^{\mathrm{\bf h}})$ (\ref*{eq:hesa}),
    $\quad$ $\mathrm{\bf y}\argt \hskip-0.5mm=\hskip-0.5mm f^{\mathrm{\bf y}}(\mathrm{\bf u}\argt, \mathrm{\bf h}\argt, \mathrm{\bf \theta}^{\mathrm{\bf y}}\argt)$, $\quad$ $\dot\w\argt \hskip-0.5mm=\hskip-0.5mm -\betaw \odot \pw\argt$ (\ref*{eq:hesb})
    \State $\ $ \vspace{-2mm}
    \State $\Hrest(\h\argt, \w\argt) = L(\y\argt, \target\argt)\phiexp\argt + \pxi\argt^{T}\dot\h\argt$ (\ref*{eq:hrestricted})
    \State $\ $ \vspace{-2mm}
\State $\dotpxi\argt = \left(\frac{\partial \Hrest(\h\argt, \w\argt)}{\partial \h\argt}\right)^{\hskip -1mm T}\hskip -1.5mm - \dis\pxi\argt$ (\ref*{eq:hesc}),
$\quad$ $\dotpw\argt = \left(\frac{\partial \Hrest(\h\argt, \w\argt)}{\partial \w\argt}\right)^{\hskip -1mm T} \hskip -1.5mm - \dis\pw\argt$ (\ref*{eq:hesd})
\State $\ $ \vspace{-2mm}
\State $\xxi\argtnext = \xxi\argt + \eulerstep \dot\h\argt$,
$\quad$ $\w\argtnext = \w\argt + \eulerstep \dot\w\argt$, $\quad$ $\pxi\argtnext = \pxi\argt + \eulerstep \dotpxi\argt$,
$\quad$ $\pw\argtnext = \pw\argt + \eulerstep\dotpw\argt$
\State $\ $ \vspace{-2mm}
\State $\t = \t + \eulerstep$
\State $\ $ \vspace{-2mm}
    \EndWhile %\vspace{-2mm} 
    %\State \hrulefill
    \end{algorithmic}
    %\vskip -2mm
    \captionof{algorithm}{Hamiltonian Learning %We show the case in which samples are evenly spaced in time (step size $\eulerstep$), but 
    (what we propose is valid also for unevenly spaced data).}
    \label{alg:hlz}
\end{algorithm} % }

\section{Hamiltonian Learning} %: Neural Computation and Learning Over Time}
\label{sec:roastedham}

%When facing the problem of learning over time from stream $\stream$, it sounds pretty natural to inherit well-established tools from optimal 
Control theory is focused on finding valid configurations (controls) to optimally drive the temporal dynamics of a system. In the case of neural networks, we aim at finding the ``optimal'' way to drive the  predictions of the model over time, by controlling the changes in $\w$. %This route is not commonly followed by existing solutions, which mostly rely on adapting gradient-based schema designed for offline learning. In this paper, we reconsider the whole learning problem to find a general solution that might open the road to novel advancements. 
This section presents HL by introducing the main involved components one after the other, and finally describing the differential equations at the core of HL. 
In order to keep HL accessible to a larger audience, we skip those formal aspects that are not explicitly needed to discuss HL from an operational perspective (see Appendix~\ref{alessandro} for a formal description of control theory).

\parafango{State-Space Formulation} Following the formalism of continuous-time state-space models \cite{gu2023mamba}, we split the computations of the network as the outcome of two sub-networks: the {\it neuron state network}, implementing function $\fxi$, is responsible of computing how $\h$ changes over time; the {\it output network}, implementing function $\fy$, is a recurrence-free portion of the network that receives data from $\h$ and, if needed, from the input units, propagating information to the output ones, as sketched in Fig.~\ref{fig:toy}.
Formally,
\begin{align}
    \dotxxi\argt = \fxi(\u\argt, \xxi\argt, \xwxi\argt) \label{eq:netsplita} & \quad\quad \text{\sc\small neuron state network}\\
    \vphantom{\dotxw}\y\argt =  \fy(\u\argt, \xxi\argt, \xwy\argt) \label{eq:netsplitc} & \quad\quad \text{\sc\small output network}\\
    \dotxw\argt =\betaw \odot \fw\argt \label{eq:netsplitb} & \quad\quad \text{\sc\small weight velocity}
\end{align}
where $\xwxi\argt$ and $\xwy\argt$ are weights of the two networks, both of them function of time, with $\xw\argt = [\xwxi\argt, \xwy\argt]$.  Eq.~\ref{eq:netsplitb} formalizes the dynamics of the weights, that will change over time.
The term $\betaw$ is a customizable vector of positive numbers of the same size of $\w$, that can be used to tune the relative importance of each weight (it can be written also as function of time), while $\fw$ is the unscaled weight velocity.
Computing $\xxi$ (resp. $\w$) requires us to integrate Eq.~\ref{eq:netsplita} (resp. Eq.~\ref{eq:netsplitb}). When using the explicit Euler's method with step $\eulerstep$, replacing the derivative with a forward finite difference, we get
\begin{equation}
    \xxi\argtnext = \xxi\argt + \eulerstep \dot\h\argt,
    \label{eq:eulera}\quad\quad\quad \w\argtnext = \w\argt + \eulerstep  \dot\w\argt, 
\end{equation}
given some $\xxi_0$ and $\w_0$. Notice that no future information is exploited, guaranteeing a form of temporal causality, and that, at time $\t$, we do not keep track of the way $\h\argt$ was computed, discarding dependencies from the past. %In Fig.~\ref{fig:nets} we show some examples of networks, emphasizing the different types of neurons we mentioned so far and showcasing some templates of connections patterns. Removing all the non-recurrent neurons and the connections involving them, we are left with one (first two examples) or more (last two examples) groups of recurrent neurons, i.e., strongly connected components of the cyclic network graph. Cycles only involve neurons from the component, and we can order components w.r.t. the signal propagation direction. When components are such that there is always a connection from any pairs of neurons belonging to each of them (in both directions), they resembles recurrent layers in classic neural networks. However, here no assumptions are made on the the way the signal propagates through components, i.e., they are not sequentially processed.

\parafango{Learning Basics} The appropriateness of the network's predictions at $\t$ is measured by a differentiable loss function $L$. It is crucial to introduce regularity on the temporal dimension, to avoid trivial solutions in which weights abruptly change from a time instant to the following one. We define {\it cost}, computed by function $C$, as a mixture of instantaneous loss $L(\y\argt, \target\argt)$, and a penalty that discourages fast changes to the weights of the model,
\begin{equation}
    C(\h\argt, \w\argt, \fw\argt, t) = L\left(\fy(\u\argt, \xxi\argt, \xwy\argt), \target\argt\right)\phi\argt + \frac{1}{2}\|\fw\argt\|^2,
    \label{eq:cost}
\end{equation}
where $\| \fw \|^{2}$ is the squared Euclidean norm of vector $\fw$.
In Eq.~\ref{eq:cost} we also introduced a (customizable) positive scalar-valued function $\phi\argt$ to tune the scale of the loss over time.
%
% Marco: qui considera che se $\phiexp$ è customizable allora $\phi$ 
% è customizable ... solo per dire che forse lo riscriverei leggermente diverso. 
% l'esponenziale è direi un buon esempio da menzionare ma non oltre.
%
The explicit dependence of $C$ on time is also due to the data (input-target) streamed at time $\t$.
If targets are not provided at time $\t$, the loss $L$ is not evaluated.

\parafango{Optimal Control} %Learning over time is framed in the context of optimal control, that is well-suited to study optimization problems on a time horizon $[\tA, \tB)$, with $N$ possibly infinite \cite{bertsekas2012dynamic}. T
The goal of learning is to minimize the {\it total cost} on the considered time horizon w.r.t. the dynamics that drive the way weights change over time, Eq.~\ref{eq:netsplitb},
\begin{equation}
    \fw \in \arg\min_{\fwtmp} \int_{\tA}^{\tB} e^{\dis\t} C(\h\argt, \w\argt, \fwtmp\argt, t)\ dt,
    \label{eq:control}
\end{equation}
which is basically saying that we are looking for the best way to move/control the weight values over time \cite{bertsekas2012dynamic}. For this reason, $\fw$ represents the optimal trajectory of the weight velocity, i.e., the optimal {\it control} of the problem.
The cost is scaled by an exponential function with $\dis \geq 0$ that has no effects when $\dis = 0$. As it will be clearer in the following, this choice introduces a form of dissipation, and we will show how to completely avoid computing the exponential. Control problems as Eq.~\ref{eq:control} are boundary value problems and must be paired with initial and final conditions, which is not the case of learning over time (no knowledge of 
 future).
%In a sense, since the agent is expected to improve over time, we require a progressively less-aggressive correction to its predictions, as it is typical in reinforcement learning {\color{red}(citation)}. %Differently, the regularizer on the weight velocities is multiplied by $e^{\dis\t}$, discouraging weight changes, since the information on the weights is expected to progressively become more stable to avoid continuous/uncontrolled changes in their configuration. 

\parafango{Costate} Let us introduce an adjoint variable named {\it costate}, $[\pxi, \pw]$, which has the same structure/dimensions of the state $[\h, \w]$. Since $\w = [\xwxi, \xwy]$, we also have $\pw = [\pwxi, \pwy]$, see Fig.~\ref{fig:toy}.
The costate is a largely diffused notion in the field of optimal control, following the Pontryagin's maximum principle \cite{bertsekas2012dynamic}, and it has a key role in our proposal, as it will be clear shortly. In a nutshell, instead of trying to directly compute the optimal way in which the weights change over time, i.e., $\fw$ of Eq.~\ref{eq:netsplitb} (that is a function valid $\forall \t$), we will redefine the control in terms of the newly introduced costate, that will make computations maneageable in our setting. Thus, we will first estimate the costate at each $\t$ and, afterwards, we will use it to evaluate $\fw$.

\parafango{Hamilton Equations} 
The way the costate changes over time and, more generally, the whole dynamics of the model are described by the so-called Hamilton Equations (HEs), which are the key component of HL. 
HEs are ODEs that provide a way of solving the problem of Eq.~\ref{eq:control}, since it can be formally shown that a minimum of Eq.~\ref{eq:control} con be reconstructed from a solution of such ODEs \cite{bertsekas2012dynamic,nostroaaai2024riccati}. %In order to bridge the cost of Eq.~\ref{eq:cost} to the HEs, we introduce  the notion of Hamiltonian $\H$, which is the sum of the (exponentially scaled) cost (the integrand of Eq.~\ref{eq:control}) and the dot product between state and costate.   
Since the HEs formalize how to compute $\dot\h$, $\dot\w$, $\dot\pxi$, $\dot\pw$ (where the first two ones are the same as Eq.~\ref{eq:netsplita}, Eq.~\ref{eq:netsplitb}), they also formally describe how weights change over time, i.e., $\dot\w$, which is what we need to let the network evolve.
HEs are computed by differentiating the so-called Hamiltonian $\H$, which is a scalar function that sums the integrand of Eq.~\ref{eq:control} with the dot product between state $[\h,\w]$ and costate $[\pxi,\pw]$. The Hamiltonian is defined in a way such that, at each $\t$, it is at a minimum value with respect to the control  (Appendix~\ref{alessandro}). Given $\H$, the HEs are $\dot\h = {\partial \H}/{\partial \pxi}$, $\dot\w = {\partial \H}/{\partial \pw}$, $\dot\pxi = -{\partial \H}/{\partial \h}$, and $\dot\pwxi = -{\partial \H}/{\partial \w}$ \cite{lewis2012optimal,betti2023neural}.

%They consist of four equations obtained by computing the gradient of $-\H$ w.r.t. the two parts of the costate variable, which yield $\dot\h$, and $\dot\w$, and the gradient of $\H$ w.r.t. to the two parts of the state variable, to get $\dot\pxi$, $\dot\pw$, respectively. Moreover, they must be paired with a constraint on the control $\fw\argt$, which must be set to a value that zeroes the gradient of the $\H$ w.r.t. it.

\parafango{Hamiltonian Learning}
In principle, the HEs at a generic time $t$ can be computed by automatic differentiation. % to get $\dot\w$, $\dot\pxi$, $\dot\pw$. 
However, HEs can only be exploited in boundary-value problems, requiring knowledge of the state at $t=0$ and of the costate at $t=\tB$, and then performing dynamic programming in the range $[\tA,\tB]$, updating the costate going backward in time, which is unfeasible when the dimension of the state and/or the length of the temporal interval is large. Since $T$ could be infinite, we have no ways of knowing $\pxi_{\tB}$ or $\pw_{\tB}$, and, as it is typical in online learning with causality constraints, we aim at tackling the learning problem in a forward manner, learning from the current state and current inputs, without having access to future information. Moreover, the exponentially scaled cost can introduce numerical issues for large exponents.
The proposed HL framework consists of a revisited instance of the Hamiltonian and of the HEs to overcome the aforementioned issues, introducing three main features highlighted in the following with ($i$.), ($ii$.), and ($iii$.). 

% Marco: forse sceglierei un altro attributo rispetto a "pragmatic"
\parafango{Forward Hamiltonian Learning} 
In the HL framework, ($i$.) we avoid to deal with an exponential function of time in $\H$, that leads to numerical issues,  by re-parameterizing the costate in an appropriate manner, yielding an exponential-free Hamiltonian $\Hrest$ and exponential-free HEs (derivations in Appendix~\ref{app:robust}). Moreover, ($ii$.) in the HEs, $\dot\w$ has a simple form which does not require automatic differentiation to be computed, thus we can avoid involving it in the computation of $\Hrest$. Putting ($i$.) and ($ii$.) together, we replace the Hamiltonian $H$ with what we refer to as {\it Robust Hamiltonian} $\Hrest$, which has no exponential functions of time and does not depend at all on the second part of the costate, $\dot\pw$,
\begin{equation}
    \Hrest(\h\argt, \w\argt) \ = L(\y\argt, \target\argt)\phiexp\argt + \pxi\argt^{T}\dot\h\argt.
    \label{eq:hrestricted}
\end{equation}
The first two HEs consist of
\def\fantasma{\vphantom{\left(\pxi\argt^T \frac{\partial \dot\h\argt}{\partial \xwxi\argt}\right)^{\hskip -1mm T}}}
\begin{align}
     \dotxxi\argt &= \fxi(\u\argt, \xxi\argt, \xwxi\argt) \tag{{\color{blue}$\mathcal{E}$$\star$1}} \label{eq:hesa}\\
    \dotxw\argt &= -\betaw \odot \pw\argt \tag{{\color{blue}$\mathcal{E}$$\star$2}} \label{eq:hesb}
\end{align}
where Eq.~\ref{eq:hesa} is the same as Eq.~\ref{eq:netsplita} (as expected), while Eq.~\ref{eq:hesb} is what we can directly compute without involving automatic differentiation, and it involves both the parameters of the state network and the ones of the output net (since $\w$ is the concatenation of both of them). The two remaining HEs are computed by differentiating the Robust Hamiltonian,
\begin{align}
   \nonumber \dotpxi\argt &= -\flipper\left(\frac{\partial \Hrest(\h\argt, \w\argt)}{\partial \h\argt}\right)^{\hskip -1mm T}\hskip -1.5mm - \dis\pxi\argt = -\flipper\left(\phiexp\argt\frac{\partial L(\y\argt,\target\argt)}{\partial \h\argt} + \pxi\argt^T \frac{\partial \dot\h\argt}{\partial \h\argt}\right)^{\hskip -1mm T} \hskip -1.5mm - \dis\pxi\argt \label{eq:hesc} \tag{{\color{red}$\mathcal{E}$$\star$3}}\\
    \nonumber \dotpw\argt &= -\flipper\left(\frac{\partial \Hrest(\h\argt, \w\argt)}{\partial \w\argt}\right)^{\hskip -1mm T} \hskip -1.5mm - \dis\pw\argt =\left[ \begin{aligned} \dotpwxi\argt \fantasma\\
    \dotpwy\argt  \fantasma
    \end{aligned} \right]  = \left[ \begin{aligned} \fantasma & - \flipper \left(\pxi\argt^T \frac{\partial \dot\h\argt}{\partial \xwxi\argt}\right)^{\hskip -1mm T} \hskip -1.5mm - \dis{\pwxi\argt} \\
   \fantasma & -\flipper\left(\phiexp\argt\frac{\partial L(\y\argt,\target\argt)}{\partial \xwy\argt}\right)^{\hskip -1mm T} \hskip -1.5mm - \dis{\pwy\argt} 
    \end{aligned} \right], \label{eq:hesd} \tag{{\color{red}$\mathcal{E}$$\star$4}}
\end{align}
where the additive terms weighed by $\dis$ are due to the re-parametrization that allowed us to remove the exponential function, and that act as dissipation terms, avoiding explosion of the dynamics. 
See Appendix~\ref{app:robust} for all the derivations.
In the case of Eq.~\ref{eq:hesd}, we also explicitly split the HE considering the contribution from the state networks and the one from the output network, respectively.
The binary value $\flipper \in \{-1, 1\}$ does not affect the dissipation terms, and is equal to $1$ in the original HEs, which are solved backward in time. %Moreover, there is bottom-to-top flow of information that is triggered by the temporal development of the four differential equations in Eq.~\ref{eq:hes}: computing $\dotxw\argt$ (which is our goal) in the $4$th equation requires $\pw\argt$, which depends on the $2$nd and $3$rd differential equations (recall that $\pw\argt = [\pwxi(t), \pwy(t)]$); in turn, the $2$nd and $3$rd equations depend on $\pxi\argt$, whose dynamics is described by the $1$st equation. 
%HEs are commonly solved using dynamic programming, updating the costate going backward in time, that is different from what is needed in the case of learning over time, where causality constraints force the model to not have access to future information, making Hamiltonian Learning not realistic. 
It has been recently shown how the HEs can be exploited to build models that work forward in time \cite{nostroaaai2024riccati}, using additional neural models to support the state-costate interaction and a time-reversed Riccati equation. Of course, the original guarantees of optimality are lost, but it opens to a viable way of learning in temporally causal manner. ($iii.$) Here we simply follow the basic intuition of reverting the direction of time when evaluating Eq.~\ref{eq:hesc} and Eq.~\ref{eq:hesd}, and solving them going forward in time. Hence, we replace the terminal condition of the boundary-value problem with an initial one, setting the costate at time $t=0$ to zeros. Then, we force $\flipper=-1$ in Eq.~\ref{eq:hesc} and Eq.~\ref{eq:hesd} to implement the aforementioned intuition about the direction of time, changing the sign of (the non-dissipative part of) the costate-related HEs.
%consider the case in which the sign of the HEs on the costate is switched, thus $\flipper=-1$ in Eq.~\ref{eq:hesc} and Eq.~\ref{eq:hesd}, and we replace the terminal condition with an initial one, i.e., we initialize the costate at time $t=0$ to zeros. 
%This choice virtually implies reverting the direction of time, and thus of the derivatives, to support costate computation {\it forward in time}. 
As we will show in Section~\ref{sec:grad}, this choice is not only driven by intuition, but it is what allows HL to correctly generalize common gradient-based learning.

\parafango{Interpretation of Hamiltonian Learning} For each $\t$, we integrate the four HEs with Euler's method, as introduced in Eq.~\ref{eq:eulera}, yielding the next values of the state and costate. We do not keep track of the past states or costates, thus HEs are local in time. This requires the network to be able to keep track of the past in order to update the learnable parameters at the current time. Exactly like the state is a form of memory of what happened so far, progressively updated over time, the costate can be interpreted as a memory of the sensitivity of the model parameters. Eq.~\ref{eq:hesc} and Eq.~\ref{eq:hesd} shows how the dynamics of the costate involve gradients with respect to the neuron outputs and parameters at time $\t$. As a consequence, when integrated, $\pxi$ and $\pw$ accumulate sensitivity information over time. We summarize HL in Algorithm~\ref{alg:hlz}, that is all that is needed to be implemented to setup HL, since derivatives can be computed by automatic differentiation.
 
%The HEs are integrated with the Euler's method, whose step is (and will be) indicated with $\eulerstep$ to keep the notation simple. Of course, such a step is not necessarily constant, depending on the properties of the stream $\stream$ and on the time that passes between consecutively processed samples, i.e., $\eulerstep\argt=\eulerstep'\Delta\argt$, where $\eulerstep'$ is a tunable scale.

\parafango{Instantaneous Propagation} 
%The continuous-time formulation presented so far represent the most generic view on the learning problem, which explicitly consider the time dynamics. 
When time does not bring any important information for the task at hand, e.g., as it is assumed in generic sequences instead of time-series, then we can cancel the effects of the selected integration procedure by a residual-like transition function in Eq.~\ref{eq:netsplita},
% Marco: $\tau^{-1}$ dovrebbe forse essere definito. E' una sorta di operatore.
% occorre non confonderlo con il passo di campionamento. 
\begin{align}
    \fxi(\u\argt, \xxi\argt, \xwxi\argt) &= \eulerstep^{-1}\left(-\xxi\argt + \fxid(\u\argt, \xxi\argt, \xwxi\argt) \right),
    \label{eq:euleraa} 
\end{align}
where the neuron state network is basically a scaled residual network. Indeed, we add $\h$ to the outcome of the neural computation implemented by the newly introduced function $\fxid$, and then we divide by $\eulerstep$.
In this way, when integrating the state equation (Eq.~\ref{eq:netsplitc}) with the Euler's method, we get $\h\argtnext =\fxid(\u\argt, \xxi\argt, \xwxi\argt)$. In other words, the next state is what is directly computed by the neural networks, discarding time and the spacing between the streamed samples. Eq.~\ref{eq:netsplitc} remains unchanged, since it is a static map. %Of course, one might reformulate Hamiltonian %Learning directly in discrete time, that is certainly doable, but that would reduce the generality of this approach, while here we present a model designed to learn over time, and not over bare sequences only.

\section{Recovering Gradient-based Learning}
\label{sec:grad}
%Marco: qui "might sound orthogonal or not clearly related" lo toglierei e andrei dritto. 
HL offers a wide perspective to model learning over time with strong locality, that will be showcased in Section~\ref{sec:good}. However, in order to help trace connections with existing mainstream technologies, in this section we neglect the important locality properties of HL to clarify the relations to the usual gradient-based minimization by BackPropagation (BP) or BackPropagation Through Time (BPTT) \cite{pascanu2013difficulty}. BPTT explicitly requires to store all the intermediate states, while BP assumes instantaneous propagation of the signal through the network. % while HL offers a larger perspective to model learning over time with strong locality, that will be showcased in Section~\ref{sec:good}.
%HL of Algorithm~\ref{alg:hlz}, at a first glance, is not clearly related to the usual gradient-based minimization of $L$ by BackPropagation (BP) or BackPropagation Through Time (BPTT) \cite{pascanu2013difficulty}. 
%However, this section shows that gradient-based learning can be recovered as a special instance of HL. %with Euler's method-based integration of the ODE, thus without requiring any software libraries or black-box functionalities to solve ODEs.
First, we show how to recover gradient-based learning in feed-forward networks (BP), presenting two different ways of modeling them in HL, either by means of the output network or of the state one. Then, we do the same for recurrent networks, where, in addition to the state/output network based implementations, we also show how HL can explicitly generalize BPTT by tweaking the way data is streamed.

%\parafango{Stochastic Sampling} Given a dataset of static examples or of sequences, we can always convert it into a continuous stream $\stream$, by providing examples over time, and repeating the process once the end of the dataset is reached. If samples are randomly sampled, then $\stream$ provides data in the order which is typical of stochastic gradient-based algorithms. In this case, the time $\t$ is the time in which an example is provided. Of course, this information is precious when $\stream$ is the virtual concatenation of sequences, where the spacing (in time) between samples of the sequences is meaningful. When it is not available or it is not important for learning purposes, we can simply assume to use a constant $\eulerstep$ or to directly rely on the discrete formulation. 
 % Marco: attenzione consistenza con operatore 

\begin{figure}
    \centering

    \includegraphics[width=0.32\linewidth,trim={2.7cm 2cm 2cm 2cm},clip]{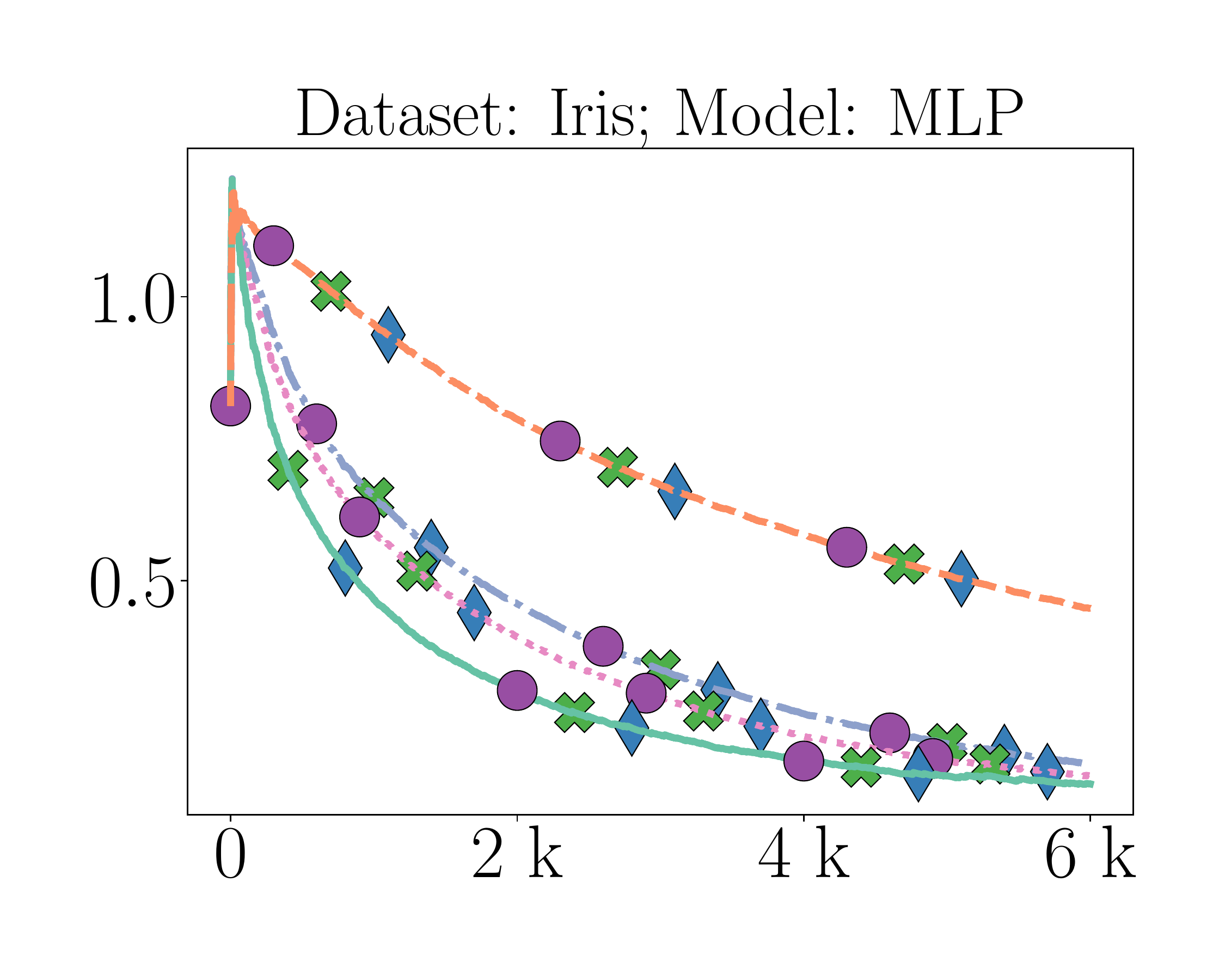} 
    \includegraphics[width=0.32\linewidth,trim={2.7cm 2cm 2cm 2cm},clip]{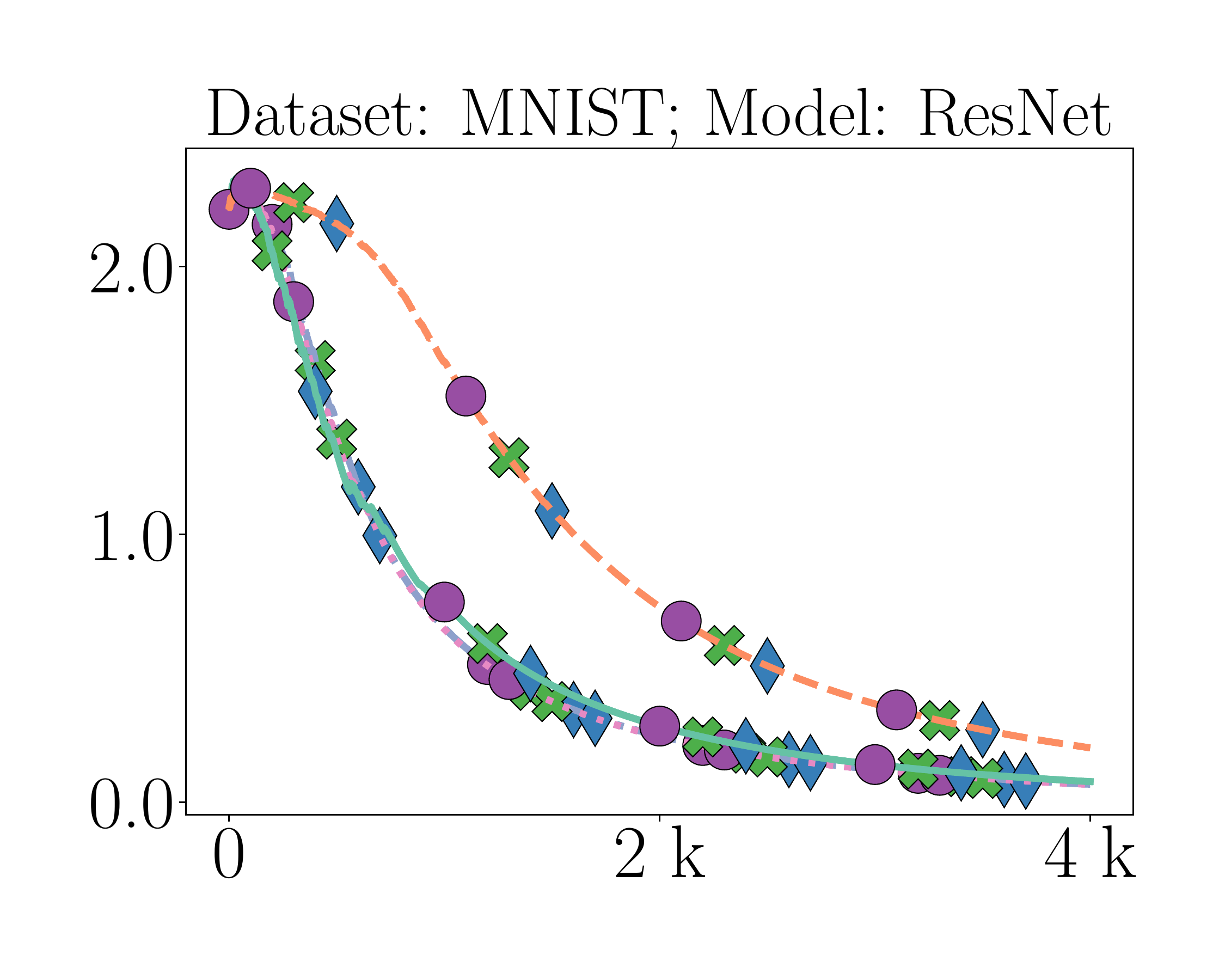}   
     \includegraphics[width=0.32\linewidth,trim={3cm 2cm 2cm 2cm},clip]{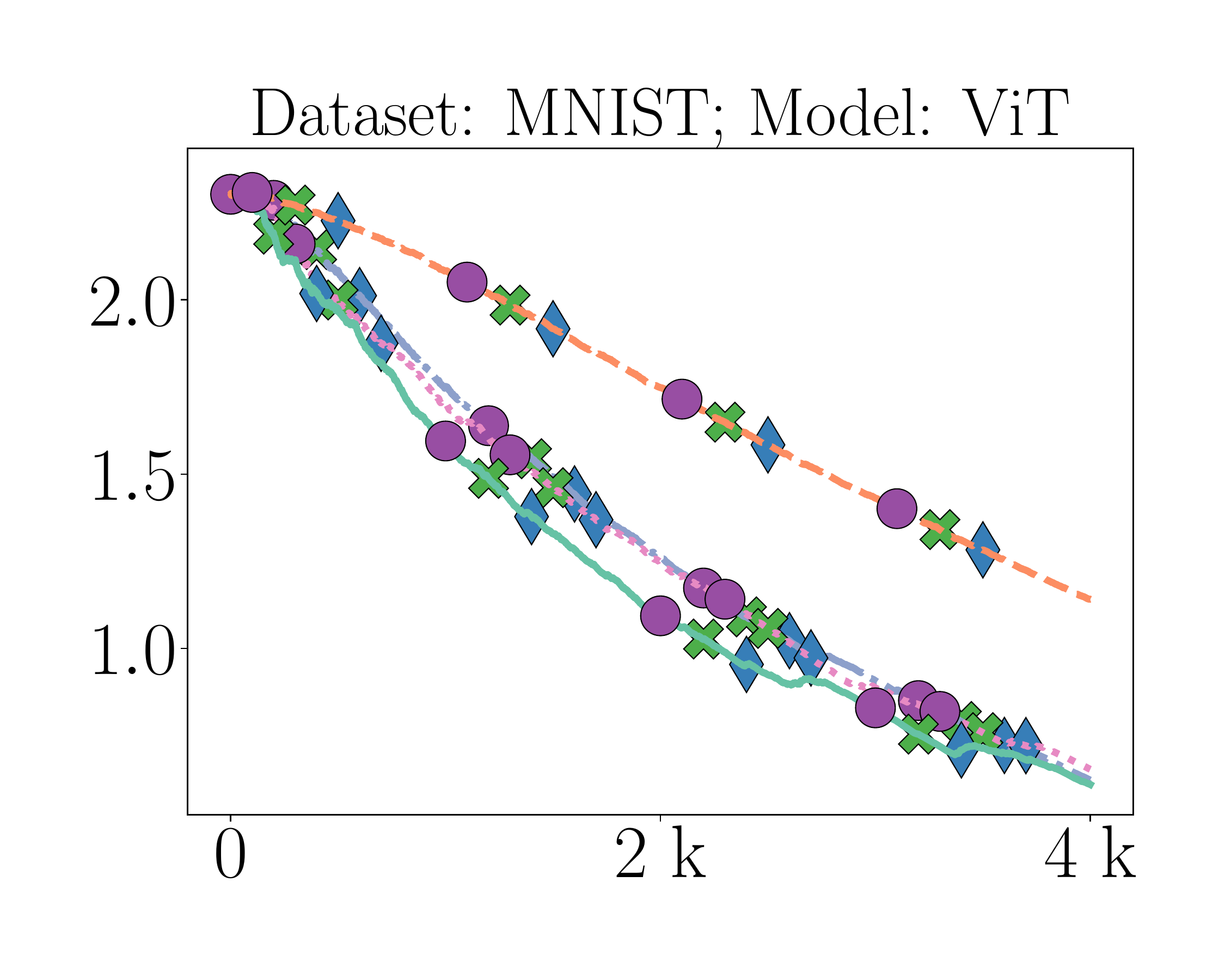}\\
    \includegraphics[width=0.32\linewidth,trim={2.7cm 2cm 2cm 2cm},clip]{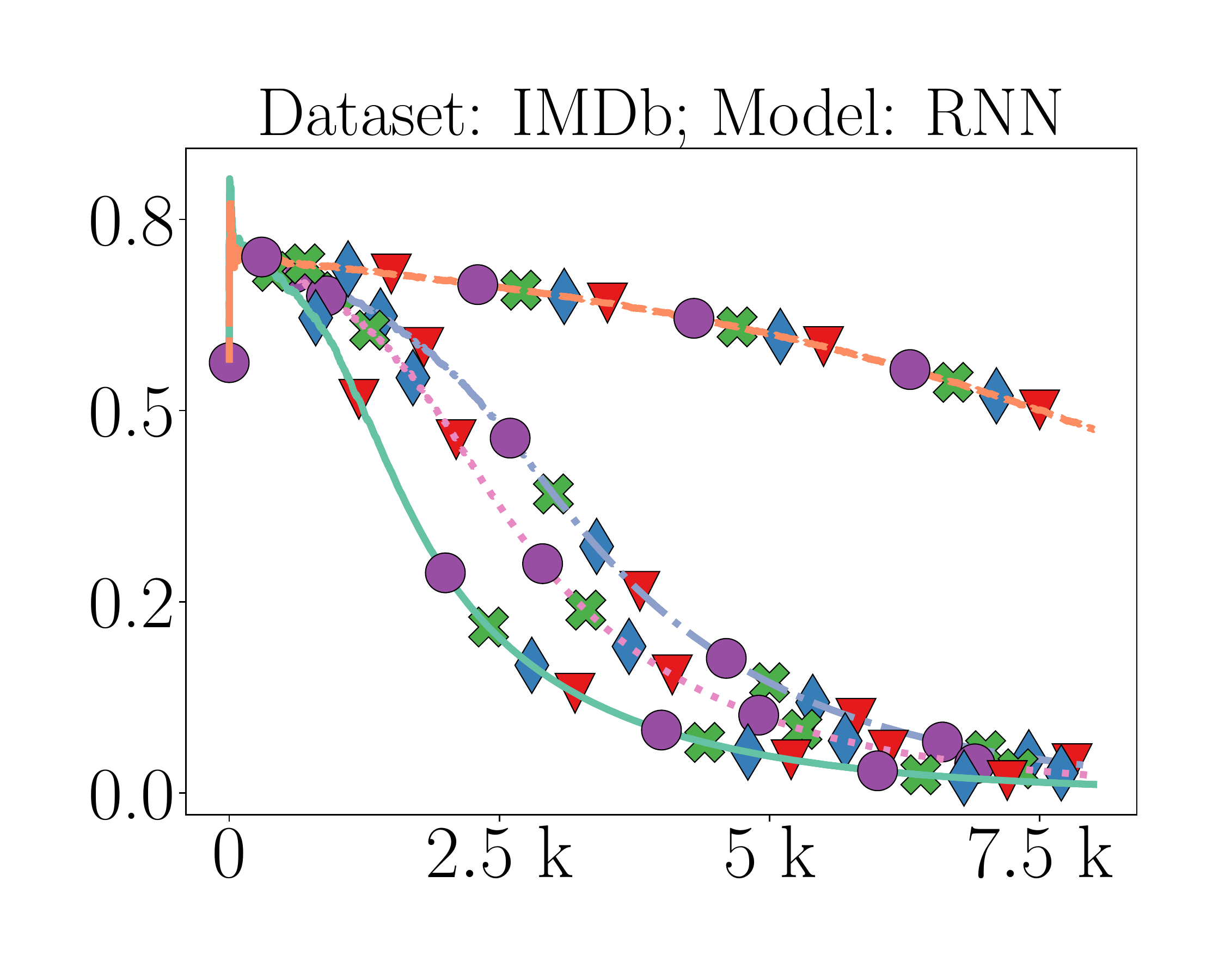}   
    \includegraphics[width=0.32\linewidth,trim={2.7cm 2cm 2cm 2cm},clip]{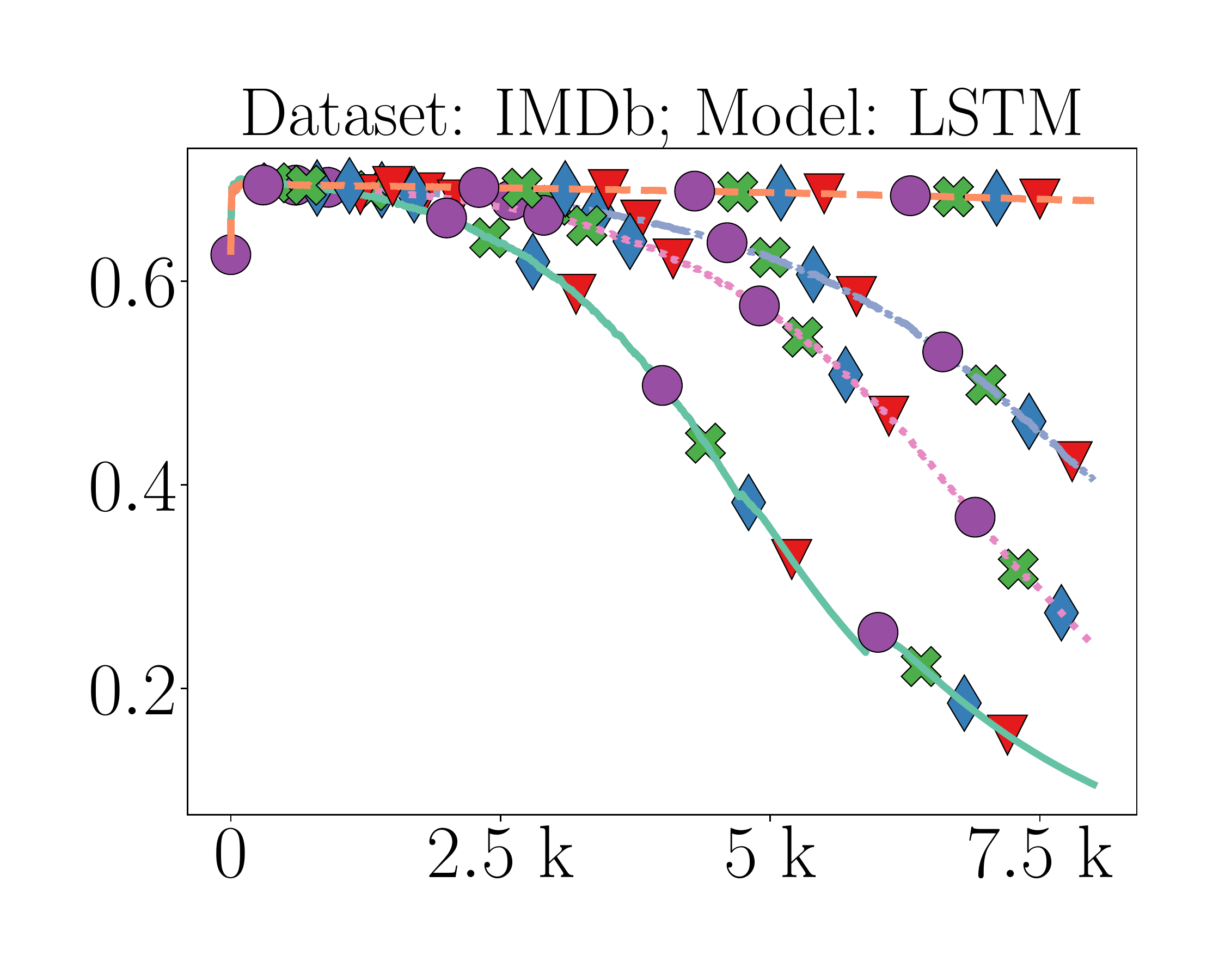} \\
    \includegraphics[width=0.7\linewidth,trim={0cm 15cm 0 0},clip]{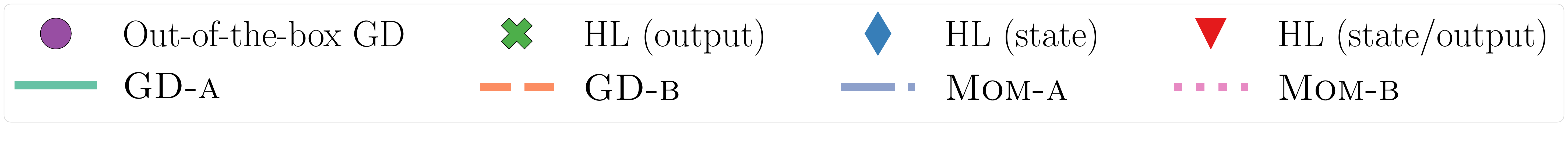}
    %\vskip -2mm
    \caption{Comparing HL with out-of-the-box Pytorch optimizers. %and Hamiltonian Learning (\textsc{HL}, setting $\tau$, $\beta$ $\dis$, $\phi$ to values that we theoretically shown to be coherent with the parameters of out-of-the-box tools-- see Appendix~\ref{oob}). 
    Each figure is about a different dataset/architecture, reporting the loss $L$ in function of time (datasets and setup in Appendix \ref{oob}); there are $4$ scenarios: w/o (\textsc{GD-a}, \textsc{GD-b}) and w/ momentum (\textsc{Mom-a}, \textsc{Mom-b}).
    In \textsc{HL}, we implemented each net by solely considering the output function ({\it output}), the state function ({\it state}), or, in the case of RNNs and LSTMs, jointly considering both ({\it state/output})--the recurrent part is implemented with the state net, the rest is the output net. %Each plot reports the value of loss $L$ in function of time. %The plot shows the perfect alignment in terms of cost function values ($y$-axis) during the models' training phase ($x$-axis, training steps).{\color{blue} va anche detto modelli di rete e datasets/dati usati/epoche -- nel main text? Appendix? fatto in appendice: } 
    Curves of each scenario have the same color/linestyle, and a per-approach marker.
    {\it The (absolute) difference between the weights yielded by the Pytorch optimizer and HL is zero in most cases or within the order of round-off error.} \bf{Thus, curves of the same scenario overlap.}}
    \label{fig:exp}
    %\vskip -5mm % UGLY!!!!
\end{figure}

\parafango{Feed-Forward Networks (Output Nets)} 
% Marco: è interessante l'analogia formale, ma si potrebbe già qui notare che con lo schema feedforward si perde lo spirito della località ... pensa a come si ricompone con le onde forward e backward
%Eq.~\ref{eq:f2} has a two-fold implementation in the state-space formulation of Hamiltonian learning, Eqs.~\ref{eq:netsplita}-\ref{eq:netsplitc}. 
Implementing a feed-forward net by means of Eq.~\ref{eq:netsplitc} sounds natural due to the static nature of the output map $\fy$. The notion of neuron state is lost, thus $\h\argt = \mathrm{\bf 0}$, $\forall t$, and, consequently
%If $\f$ of Eq.~\ref{eq:f2} is implemented by the output function $\fy$, then there is no state network (thus no Eq.~\ref{eq:netsplita}, Eq.~\ref{eq:eulera}, Eq.~\ref{eq:hesa}, Eq.~\ref{eq:hesc}), which sounds natural once we recall that Eq.~\ref{eq:f2} is actually independent from $\bar\h$, being it an instantaneous map. 
$\Hrest$ of Eq.~\ref{eq:hrestricted} boils down to the usual loss function. The costate-related HE of Eq.~\ref{eq:hesd} (lower part), due to our choice of $s=-1$, is the gradient of the loss with respect to the weights  (let us temporarily discard the dissipation term). When integrating the output-net-related HEs, Eq.~\ref{eq:hesd}, Eq.~\ref{eq:hesb}, respectively, setting $\Hrest = L$ and $s=-1$, we get
\begin{align}
    \pwy\argtnext &= \pwy\argt + \eulerstep\left(\frac{\partial L(\y\argt, \target\argt)}{\partial \xwy\argt}\phiexp\argt\right)^{\hskip -1mm T} \hskip -1.5mm - \eulerstep\dis\pwy\argt = (1 - \eulerstep\dis)\pwy\argt + \eulerstep\phiexp\argt\left(\frac{\partial L(\y\argt, \target\argt)}{\partial \w\argt}\right)^{\hskip -1mm T} \hskip -1.5mm  \label{eq:fanculo} \\
    \xwy\argtnext &= \xwy\argt - \eulerstep\betawy\pwy\argt.\label{eq:fanculo2}
\end{align}
%We wrote $L\argt$ in place of $L(\y\argt, \target\argt)$, to simplify the notation.
%it is evident how such equations consist in a momentum-based computation of the gradient of $L$ with respect to $\xwy$  \cite{sutskever2013importance}, and a weight update step, respectively,
Eq.~\ref{eq:fanculo} is a momentum-based computation of the gradient of $L$ with respect to $\xwy$  \cite{sutskever2013importance}, where $1-\eulerstep\dis$ is the momentum factor and  $\eulerstep\phiexp\argt$ is a time-dependent dampening coefficient. Eq.~\ref{eq:fanculo2} is the classic weight update step, as the one we introduced in Eq.~\ref{eq:wwww}, with (per-parameter) learning rate $\eulerstep\betawy$, being $\betawy$ the function $\betaw$ restricted to the parameters of the output network. However, in classic gradient-based learning, a specific order of operations is strictly followed: first gradient computation and, only afterwards, weights are updated with the just computed gradients. Hence, in HL we have to replace $\pwy\argt$ of Eq.~\ref{eq:fanculo2} with $\pwy\argtnext$, and then use the definition of $\pwy\argtnext$ given by Eq.~\ref{eq:fanculo}. Under this constraint, we get a perfect match with classic gradient-based learning. Notice that when $\dis = 1/\eulerstep$, the momentum term disappears. In such a setting, if $\phiexp$ is an exponential function, it acts as an exponential scheduler of the learning rate, while if $\phiexp$ is constant then no scheduling is applied \cite{Li2020An}. % In the discrete case, we get an analogous result, with the only differences being that the learning rate is $\betawy$, and the momentum factor $\dis$ (thus it disappears if $\dis = 0$). We mention that, in both the continuous and discrete cases, 
Of course, we can also get rid of the momentum term by zeroing the weight-costate, $\pwy\argt = \mathrm{\bf 0}$  right before processing the example at time $\t$. In Appendix~\ref{oob} we report the exact rules to map learning parameters in momentum based gradient-descent and HL parameters. Fig.~\ref{fig:exp} reports the outcome of comparing gradient-based learning (with and without momentum) using popular out-of-the-box tools (different learning rates, momentum terms, damping factors and networks) and HL, showing that they lead to the same trajectories of the loss function, considering MLPs, Transformers (ViT), ResNets (see also Appendix~\ref{oob}).

\parafango{Feed-Forward Networks (State Nets)} Implementing a feed-forward network  using the state network requires to clear the state $\h\argt$ and costate $\pxi\argt$ for each $\t$, setting them to $\mathrm{\bf 0}$, to avoid propagating neuron-level information from previous examples. In this case, $\fy$ is the identity function, %Since $\f$ of Eq.~\ref{eq:f2} is an instantaneous map, whose output does not depend on the past, right before evaluating the state transition and the HEs, we clear the neuron state and costate for each $\t$, i.e.,  $\xi\argt = \mathrm{\bf 0}$ and $\pxi\argt = \mathrm{\bf 0}$
and we can exploit the residual-like formulation of Eq.~\ref{eq:euleraa} to compensate the effects of the integration procedure (that depends on $\eulerstep$). In such a setting, we get (details in Appendix~\ref{ffandstate})
%\begin{equation}
%    \h\argtnext = \eulerstep\fxi(\u\argt, \mathrm{\bf 0}, \xwxi\argt) / \eulerstep = \dot\h\argt,
%    \quad \pxi\argtnext = -\eulerstep\flipper\phiexp\argt\frac{\partial L(\h\argtnext,\target\argt)}{\eulerstep\partial \h\argtnext} = \dotpxi\argt \label{eq:figlio}.
%\end{equation}
\begin{align}
    \h\argtnext &= \fxid(\u\argt, \h\argt, \xwxi\argt)= \eulerstep\dot\h\argt, \label{eq:figliotop}\\ \pxi\argtnext^T &= -\eulerstep\flipper\phiexp\argt\frac{\partial L(\h\argtnext,\target\argt)}{\partial \h\argtnext}, %= \eulerstep\dotpxi\argt
    \label{eq:figliobottom}
\end{align}
where we used $\h\argtnext$ as argument of $L$ instead of $\h\argt$. This choice is the direct counterpart of what we did when enforcing sequential ordering in gradient computation and weight update in the previously discussed implementation.
In fact, we want the evaluation of the loss $L$ to happen after the state neurons have been updated ($\h\argtnext$), which is coherent with what happens when performing a prediction first, and only afterward evaluating the loss. Notice that, for the same reason, we also differentiate with respect to $\h\argtnext$. %This choice is the direct counterpart of what we did when enforcing sequential ordering in gradient computation and weight update when discussing the previous  implementation. 
Eq.~\ref{eq:figliotop} and Eq.~\ref{eq:figliobottom} collect the two HEs of  Eq.~\ref{eq:hesa} and Eq.~\ref{eq:hesc}, respectively. In turn, the HE of Eq.~\ref{eq:hesd} becomes,
\begin{equation}
\begin{split}
    \nonumber \dotpwxi\argt &=  - \flipper \left( \pxi\argtnext^T \frac{\partial \dot\h\argt}{\partial \xwxi\argt}\right)^{\hskip -1mm T} \hskip -2mm -  \hskip -0.5mm \dis\pwxi\argt
    = \hskip -0.5mm \left(\phiexp\argt\frac{\partial L(\h\argtnext,\target\argt)}{\partial \h\argtnext} \frac{\partial \h\argtnext}{\partial \xwxi\argt}\right)^{\hskip -1mm T}  - \dis\pwxi\argt \\
    & = \left(\phiexp\argt\frac{\partial L(\h\argtnext,\target\argt)}{\partial \xwxi\argt}\right)^{\hskip -1mm T} \hskip -1.5mm - \dis\pwxi\argt,
\end{split}
\label{eq:hesdbla}
\end{equation}
where, in the first equality, we replaced $\pxi\argtnext$ with its definition reported in Eq.~\ref{eq:figliobottom}, while in the second one we replaced $\dot\h\argt$ with $\tau^{-1}\h\argtnext$ accordingly to Eq.~\ref{eq:figliotop}, and discarded $\flipper$, since $\flipper^2 = 1$ (that is expected, since we are collapsing the dynamics of the two costate HEs). The last equality holds due to the chain rule. When integrated, it yields the same equation that we discussed in the previous implementation, Eq.~\ref{eq:fanculo}.\footnote{Of course, replacing $\xwy$ with $\xwxi$ and $\pwy$ with $\pwxi$.} Then, Eq.~\ref{eq:fanculo2} still holds also in this case. As a consequence, all the comments we made about the analogies with momentum-based gradient learning, BP, and learning rate scheduling are valid also in this case.
See Fig.~\ref{fig:exp} for comparisons with plain gradient-descent (MLP, ViT, ResNet).

\parafango{Recurrent Neural Networks (Automatic Unfolding)} 
%Marco: BPTT mi pare si ritrova solo se si considera unfolding di un passo. 
Given a sequence of $n$ samples, streamed at times $t_0, \ldots t_{n-1}$, learning in Recurrent Neural Nets (RNNs) is driven by BPTT \cite{pascanu2013difficulty}, where $n$ states are stored while going forward, with constant weights $\w$, and then gradients of $\sum_{i=0}^{n-1}L(\y_{\t_i}, \target_{\t_i})$ with respect to $\w$ are computed in a backward manner. %While this seems in contrast with the forward-only nature of Hamiltonian Learning, we can recover BPTT in a two-fold manner.
% ONE
The most straightforward way to trace a connection between BPTT and HL is by conceiving the recurrent network in its unfolded form,  and by rethinking $\stream$. If we assume $\stream$ to yield one-full sequence at each time step, then we can switch our attention from the RNN to the feed-forward network obtained by unfolding it, that receives as input the whole sequence at once. Being it a feed-forward network, it can be implemented as discussed in the above text. As usual, the unfolding can be done on the fly, thus either $\fxi$ or $\fy$ can be specifically implemented as RNNs, reusing existing software implementations. This is what we did in the Movie Review experiments (IMDb \cite{imdb} data) of Fig.~\ref{fig:exp}.

\parafango{Recurrent Neural Networks (Hamiltonian Learning)} 
A more intriguing way to recover BPTT is obtained by keeping our original definition of $\stream$, which yields a sequence streamed one-token-per-time ($n$ tokens), from $\t_0$ to $\t_{n-1}$, followed by the same tokens in reverse order (without repeating the one at time $\t_{n-1}$, i.e., the sample at  $\t_{n-1+j}$ is equal to the one at $\t_{n-1-j}$, $\forall j \geq 0$).
We still exploit an identity output function and the formulation of Eq.~\ref{eq:euleraa} to compensate the effects of the integration. Of course we have to enforce the processing of the streamed data with the same weights, and only afterwards compute their variation, that can be achieved forcing 
%We manipulate $\stream$ to yield the sequence in reverse order right after time $t_{n-1}$ (assuming sample $t_{n}$ to be a blank pivot and $t_{n+j} = t_{n-j}$, $\forall j > 0$). Recall that $\stream$ yields $\delta({t_{n-1}}) = 1$ to signal that $\u(t_{n-1})$ is the last element of the sequence. To simplify the description, let us assume that all samples are evenly spaced by $\eulerstep$. 
$\betaw = \mathrm{\bf 0}$, $\forall \t \neq 2\t_{n}-\t_0$, which is indeed a degenerate condition for $\betaw$ (see Eq.~\ref{eq:hesb}). When $\t \leq \t_{n-1}$, given some initial $\h_{t_0}$, $\w_{t_0}$, the first two HEs, Eq.~\ref{eq:hesa}, Eq.~\ref{eq:hesb}, will drive the evolution of the system.
Coherently with BPTT, we must store the sequence of the $n$ generated $\h$'s, loosing locality. The last two HEs, Eq.~\ref{eq:hesc}, Eq.~\ref{eq:hesd}, do not bring any information, since changes in the costate will not affect $\w$, due to our choice of zeroing $\betaw$. When $t = t_{n_1}$, all the HEs play an important role, while when $\t > \t_{n-1}$, we basically swap the relevance of the two pairs of HEs, giving emphasis to Eq.~\ref{eq:hesc} and Eq.~\ref{eq:hesd}, and ignoring Eq.~\ref{eq:hesa} and Eq.~\ref{eq:hesb}, since we already stored the $n$ states. This is motivated by fact that we are going to process the reversed sequence, and we do care about tracking the sensitivity of the model with respect to changes in the learnable parameters, which is the role of the costate. The sequence of neuron states was already pre-computed, and it can be retrieved by mapping time by $\psi(\t) \colon \t \to 2\t_{n-1}-2\eulerstep-\t$, for $\t > \t_{n-1}$. %The costate at $\t = \t_{n-1}$ must be set to $\mathrm{\bf 0}$, to clear past information. 
When integrating the costate equations, using the proposed $s=-1$, we get (details in Appendix~\ref{bptt}),
%\begin{align}
%   \pxi\argtnext &= \pxi\argt + \eulerstep\left(\phiexp\argt\frac{\partial L(\h\argpsit,\target(\psi(\t)-\eulerstep)}{\partial \h\argpsit} -  \pxi\argt \frac{\partial \dot\h\argpsit}{\partial \h\argpsit} + \dis\pxi\argt \right) \label{eq:hescbptt} \\
% \pwxi\argtnext  &=  \pwxi\argt +\eulerstep \pxi\argt \frac{\partial \dot\h\argpsit}{\partial \xwxi\argpsit} + \eulerstep\dis\pwxi\argt \label{eq:hesdbptt}.
%\end{align}
\iffalse
\begin{align}
   \pxi\argtnext^T &= \eulerstep\phiexp\argt\frac{\partial L(\h\argpsit,\target\argpsitnext)}{\partial \h\argpsit} + \pxi\argt^T \frac{\partial \h\argpsitprev}{\partial \h\argpsit} - \eulerstep\dis\pxi\argt^T \label{eq:hescbptt} \\
 \nonumber \pwxi\argtnext  &=  \pwxi\argt +\eulerstep \left(\pxi\argtnext^T \frac{\partial \dot\h\argpsitnext}{\partial \xwxi\argpsit}\right)^{\hskip -1mm T} \hskip -1.5mm - \eulerstep\dis\pwxi\argt\\ 
 &= \pwxi\argt + \left(\pxi\argtnext^T \frac{\partial \h\argpsit}{\partial \xwxi\argpsit}\right)^{\hskip -1mm T} \hskip -1.5mm - \eulerstep\dis\pwxi\argt  \label{eq:hesdbptt}.
\end{align}
\fi
\begin{align}
   \pxi\argtnext^T &= \eulerstep\phiexp\argt\frac{\partial L(\h\argtnext,\target\argt)}{\partial \h\argtnext} + \pxi\argt^T \frac{\partial \h_{\t+2\eulerstep}}{\partial \h\argtnext} - \eulerstep\dis\pxi\argt^T \label{eq:hescbpttx} \\
 \nonumber \pwxi\argtnext  &=  \pwxi\argt +\eulerstep \left(\pxi\argtnext^T \frac{\partial \dot\h\argt}{\partial \xwxi\argt}\right)^{\hskip -1mm T} \hskip -1.5mm - \eulerstep\dis\pwxi\argt\\ 
 &= \pwxi\argt + \left(\pxi\argtnext^T \frac{\partial \h\argtnext}{\partial \xwxi\argt}\right)^{\hskip -1mm T} \hskip -1.5mm - \eulerstep\dis\pwxi\argt  \label{eq:hesdbpttx}.
\end{align}
%The same equation is responsible of vanishing the effects of the selected integration strategy when updating the state. 
where in the HE of Eq.~\ref{eq:hesc}, coherently with what we did for feed-forward networks, we anticipated the state-update, thus all the terms involving $\h\argt$ or $\dot\h\argt$ are replaced with $\h\argtnext$ and $\dot\h\argtnext$, respectively. 
In order to match BPTT, at $\t = \t_{n-1}$ we set the costate to $\mathrm{\bf 0}$ to clear past information (notice that while $\pxi$ must be set to zeros, we can also avoid resetting $\pwxi$, that will inherently introduce momentum, as already discussed), and we replace $\t$ with $\psi\argt -\eulerstep$ in the time index of non-costate-related quantities, yielding 
\begin{align}
    \hskip -2.5mm\pxi\argtnext^T  &= \eulerstep\phiexp_{{\psi\argt - \eulerstep}}\frac{\partial L(\h\argpsit,\target_{\psi\argt-\eulerstep})}{\partial \h\argpsit} + \pxi\argt^T \frac{\partial \h_{\psi\argt+\eulerstep}}{\partial \h\argpsit} - \hskip -0.5mm \eulerstep\dis\pxi\argt^T \label{eq:hescbptt} \\
 \hskip -2.5mm\pwxi\argtnext   & = \pwxi\argt + \left(\pxi\argtnext^T \frac{\partial \h\argpsit}{\partial \xwxi_{\psi\argt - \eulerstep}}\right)^{\hskip -1mm T} \hskip -2mm - \hskip -0.5mm \eulerstep\dis\pwxi\argt  \label{eq:hesdbptt}.
\end{align}
When integrating the costate, the step size $\eulerstep$ still has a role. If we select $\eulerstep = 1$ and $\dis = 0$ (no dissipation), or $\phiexp\argt = \eulerstep^{-1}$ and $\dis = 0$, Eq.~\ref{eq:hescbptt} and Eq.~\ref{eq:hesdbptt} are the largely known update rules of BPTT for sequential data, as detailed in Appendix~\ref{bptt}. %In fact, discarding the dissipation terms weighed by $\dis$, Eq.~\ref{eq:hescbptt} is equivalent to Eq.~\ref{eq:bptt1} in Appendix~\ref{bptt}, while Eq.~\ref{eq:hesdbptt} is Eq.~\ref{eq:bptt2} in Appendix~\ref{bptt}, which is the gradient of the loss function w.r.t. to the learnable parameters. 
This correspondence has the important role of confirming the validity on the choice we made in setting $s=-1$ in HL.

\parafango{Recurrent Neural Networks (Truncated)} Truncated BPTT of length $r$ can be recovered by only resetting the state at the very beginning of the input sequence and, once the end of the sequence has been reached, by streaming the last $r-1$ tokens in reverse order, going back to the previously discussed case (which implies setting neuron-costate to zero right before processing the last token of the original sequence). If $r=1$, we match the case of feed-forward neural networks, even if the initial state is not zeroed/ignored.

% - MNIST http://yann.lecun.com/exdb/mnist/
% - Iris This dataset is licensed under a Creative Commons Attribution 4.0 International (CC BY 4.0) license. https://archive.ics.uci.edu/dataset/53/iris
% - IMDb https://ai.stanford.edu/~amaas/data/sentiment/ (\cite{imdb}
% Experiments executed on a Linux server with Pytorch 2.3.0, equipped with a 10th-Gen Intel CPU and 64GB of RAM (experimens take minutes to run, log, and plot results).

\section{Leveraging Hamiltonian Learning}
\label{sec:good}
HL is designed to drive learning over time in {\it stateful} models in a temporally local manner. While it is general enough to recover the popular gradient-based learning in feed-forward and recurrent nets, as we discussed in Section~\ref{sec:grad}, the importance of dealing with stateful networks goes beyond it. We distinguish among two use-cases where HL can be leveraged to design efficient models that learn over time.

\parafango{Fully Local Learning}
%Hamiltonian Learning is temporally local by design. In fact, the HEs, Eq.~\ref{eq:hesa}-\ref{eq:hesd}, update the model state/costate only considering the previous state/costate. 
The stateful nature of learning allows us to instantiate networks that are not only temporally local, but also spatially local, once we design $\fxi$ to return the variations of the output values of {\it all} the its neurons. %, i.e., the state $\h$ is the set of outputs of all the neurons in the network. 
Given the graph that describes the parent/child relations among neurons, being it Directed Acyclic/Cyclic Graph, all the neurons can compute their outputs in parallel. In fact, they have the use of the full state $\h\argt$ (and/or the current $\u\argt$)--computed during the previous time step--to feed their inputs, with important connections to biologically plausible computational models and related studies \cite{stork,sacramento2018dendritic,meulemans2021credit}. This choice introduces delays in the computations, but it yields fully local learning, with known benefits in terms of parallelization \cite{marra2020local} and pipeline parallelism in neural networks for multi-device computation \cite{meloni} (Appendix~\ref{further}). %The idea of spatially local networks (with delays) has been studied in previous works---see \cite{marra2020local} and references therein---usually introducing specific constraints and adjoint variables to allow neurons to operate in parallel. In the case of HL, spatial locality is naturally achieved by the aforementioned definition of state. % and computational graph, going beyond the case of DAGs which is commonly discussed in the scientific literature.
%When updating the costate, spatial locality implies that the Jacobians in Eq.~\ref{eq:hesc} and Eq.~\ref{eq:hesd} will be non-zero only in positions associated to pairs of neurons with are directly connected and in positions associated to weight-neuron pairs where the weight directly enters the neuron. Hence, also these equations are spatially local.

%\parafango{Distributed Neural Computations}
%The aforementioned notion of spatial locality can be structured in several different ways. %Given the DAG associated to a In a multi-layer feed-forward network, we can create groups of consecutive layers, and assume $\fxi$ returns the variation of the output values of the {\it last layer of each group}, thus the state $\h$ is not composed by all the neuron outputs of the net. Computations within each group happens instantaneously, and delays are only among groups. This weakly spatially local architecture corresponds to the one exploited to build pipeline parallelism in neural networks \cite{meloni}, when each group is processed on a different devices/GPUs. Interestingly, the notion of costate in HL is what allows the transfer of learning-related information  among groups, that in \cite{meloni} is stored into appositely introduced variables.

\parafango{Memory Efficient BackPropagation}
%Throughout the whole paper we assumed to integrate ODEs with the Euler's method, mostly for simplicity, and also to help trace connections with gradient methods. 
The popular Neural ODEs/CDEs \cite{neuralode,neuralcde} exploit a stateful model to implement the way the signal propagates through a feed-forward network \cite{neuralode} or when providing fixed-length sequences \cite{neuralcde}, using an ODE solver to update the state and, afterwards, to compute gradients. These models do not require to store the output of all the ``layers'' (which is a vague concept in Neural ODE/CDEs) while backpropagating the learning signal, since the second call to the ODE solver can basically estimate the output of the previous layer given the output of the following one. This property holds in the continuous formulation of the models, which might be prone to numerical errors when integrating on a number of discrete steps \cite{zhu2022numerical,anode,anode2}.\footnote{Costate in HL corresponds to adjoint sensitivity in Neural ODE, being them inherited from the same tools of control theory.}
Similarly to Neural ODE, in HL, given $\u$ (generic input sample--we removed the time index), we can distribute the sequential computation of layers over time (see Appendix~\ref{further}, Eq.~\ref{eq:ffasd}). If we integrate the state-transition with the midpoint rule \cite{zhu2022numerical}, Eq.~\ref{eq:eulera} becomes
\begin{equation}
    \h\argtnext = \h\argtprev + \tau {\fxi}(\u, \h\argt, \xwxi\argt),
    \label{eq:mid}
\end{equation}
that, given $\h\argtprev$ and $\h\argt$ (we discard the other arguments of $\fxi$ in this discussion), allows us to  compute $\h\argtnext$. However, the direction of the integration can be also reversed, and the midpoint rule can also yield the previous state given the following one(s), i.e., we can compute $\h\argtprev$ given $\h\argt$ and $\h\argtnext$, by $\h\argtprev = \h\argtnext - \tau {\fxi}(\u, \h\argt, \xwxi\argt)$. Notice that in both the case (direct or reversed), function $\fxi$ is evaluated on the same point (differently from what would happen with Euler's method). 
%This opens to BackPropagation without storing the outputs/activations of all the layers, since, outputs. 
When using the midpoint rule, Eq.~\ref{eq:mid}, HL does not have any numerical issues in re-building the output of the previous layer given the output of the following one.
Moreover, we observe that two consecutive integration steps in HL, $\h\argtnext = \h\argtprev + \tau {\fxi}(\u, \h\argt, \xwxi\argt)$ and $\h_{\t+2\eulerstep} = \h\argt + \tau {\fxi}(\u, \h\argtnext, \xwxi\argtnext)$ are exactly the equations that drive the notion of Reversible ReLU
\cite{gomez2017reversible}, once we replace $y_1$, $y_2$, $x_1$, $x_2$ of Eq. (6) in \cite{gomez2017reversible}, with $\h\argtnext$, $\h_{\t+2\eulerstep}$, $\h\argtprev$, $\h\argt$, respectively, and assuming $\mathcal{F} = \mathcal{G} = \fxi$, where $\mathcal{F}, \mathcal{G}$ are the transition functions of Eq. (6) in \cite{gomez2017reversible}. Reversible ReLU allows to compute gradients without storing activations, and they are based on specifically engineered blocks. Interestingly, in HL, the same properties are simply the outcome of having changed the integration technique.

\parafango{Limitations}
%\label{sec:limitations}
HL is a way to approach learning over time in a principled manner, % that generalizes classic online gradient, but this
but, of course, it does not solve the usual issues of continual learning (e.g., catastrophic forgetting \cite{wang2024comprehensive}). Due to its forward-over-time nature, it is well-suited to deal with streams of perceptual data that does not change too abruptly. In fact, in the most local formulation of HL (Section~\ref{sec:good}), there are intrinsic delays in the propagation of information, since state and costate are updated at the same time, and weights keep changing while handing the streamed data %As a result, if the data incurs in frequent and fast modifications over time, then this might create issues to the learning process 
(see Appendix~\ref{further} for discussions on how to trade-off locality in the computations and delays).

\section{Related Work}
\label{sec:related}
\parafango{Optimal Control}
Optimal control theory \cite{lewis2012optimal} provides a clear framework to handle optimization problems on a temporal horizon (Pontryagin Maximum Principle \cite{gamkrelidze1964mathematical,giaquinta2013calculus}, Dynamic Programming \cite{bardi1997optimal}), usually involving boundary-value problems which require iterative forward/backward schemes over the whole considered time interval or working with receding horizon control~\cite{garcia1989model}. \citet{Jin2019PontryaginDP} considered the possibility of exploiting HEs for learning system dynamics and controlling policies forward-in-time, while \cite{betti2022continual}
evaluated HEs in continual online learning, artificially forcing the costate dynamic to converge to zero. %In this work, Hamilton Equations are used for driving the dynamics of an optimization problem involving the temporal variations of the weights. The proposed initial-value problem with forward-in-time solution is general enough to recover common gradient descent optimization in neural networks. 
To our best knowledge, the relations between HEs, control theory, and gradient-based approaches were not studied from a foundational perspective in the context of learning from a stream of data with neural networks. The work of \citet{lecun} mentions them, indeed, in relation with classical mechanics.

\parafango{Online Learning} There exist several works focussed on learning in an online manner from streamed data, such as in the case of physics-inspired models \cite{betti2019cognitive,tiezzi2020focus} or approaches to continual online learning \cite{MAI202228,wang2024comprehensive}. The framework of this paper is devised by revisiting the learning problem from scratch, instead of trying to directly adapt classic statistical approaches typical of offline learning.

\parafango{Neural ODE} The methods exploited in this paper are inherited from optimal control, as it is done also in the case of Neural ODE and related works \cite{neuralode,neuralcde,anode,anode2,massaroli2020dissecting}. However, here we focus on the problem of learning over time in an online manner, with a possibly infinite horizon, which is different from what is commonly done in the literature of Neural ODE/CDE. 

\parafango{Real-Time Recurrent Learning} The classic approach to learning online with recurrent models is RTRL \cite{irie2023exploring}, which requires to store, exploit, and progressively update the temporal Jacobian matrix, with high space/time complexities. Several approximations were proposed to reduce the complexity of RTRL (UORO and others, see \cite{marschall2020unified}). In HL, no temporal Jacobian matrices are stored. Hence, it is not a generalization of RTRL and related work (such as Online LRU \cite{zucchet2023online}).

\parafango{Brain-Inspired Computing} Neural approaches that are brain-inspired (to different extents), such as
predictive coding
\cite{salvatori2023brain,rao1999predictive,millidge2022predictive}, target propagation, local representation alignment
\cite{ororbia2019biologically,lee2015difference}, forward-only learning \cite{kohan2018error,ororbia2023predictive,hinton2022forward}, share the principles of locality in the computations which has been described in Section~\ref{sec:good}. HL is not inspired by brain-related dynamics but by the idea of learning over time. %A direction of future work could be framing brain-inspired models in the context of HL of this paper. %, which does not pose restrictions on the form of the state-transition function or the one of loss function.

%\vskip -10mm

\section{Conclusions}
\label{sec:conclusions}
We presented HL, a unified framework for neural computation and learning over time, exploiting tools from control theory. Differential equations drive learning, leveraging stateful networks that are fully local in time and space, recovering classic BackPropagation 
 (and BPTT) in the case of non-local models. HL represents a novel perspective, that might inspire researchers to further investigate learning over time in a principled manner. %, since we just scratched the surface of what could be possibly done. 
%In future work, we plan to focus on neural architectures that, when paired with Hamiltonian Learning, favour memorization skills, promoting generalization and reducing forgetting.
%Due to the foundational aspects of what is proposed in this paper, there are no potential negative societal impacts we think should be mentioned.

%\bibliographystyle{unsrtnat}
%\bibliography{biblio}

\appendix
\section{Further Details}
\label{further}

\parafango{Stream} The perspective of this paper is the one in which learning consists of the online processing of a single stream of data, that could be possibly infinite. 
In order to convert the classic notion of dataset to the one of stream $\stream$, we can consider that $\stream$, at time $\t$, yields an input-target-tag triple, $\stream\argt = (\u\argt, \target\argt, \delta\argt)$, being $\delta\argt$ a binary tag whose role will be clear in the following.
A given dataset of samples can be streamed one sample after the other, possibly randomizing the order in case of stochastic learning, and $\delta\argt=1$, $\forall \t$. If samples are sequences, then they are streamed one after the other. However, in this case each single sequence must be further streamed token-by-token, thus $\stream$ provides a triple with $\delta\argt = 1$ for the last token of the current sequence (otherwise $\delta\argt=0$), to preserve the information about the boundary between consecutively streamed sequences.
%The source of information is a stream $\stream$ that, at time $\t$, yields an input-target-tag triple, $\stream\argt = (\u\argt, \target\argt, \delta\argt)$,\footnote{Targets could also be present only for some $\t$'s.} where the binary tag $\delta$ is meaningful only if the stream is segmented, and it is $1$ for the last batch of a segment. 
%For example, a dataset of sequences becomes a unique long sequence composed of the concatenation of the contents of the dataset, which is progressively streamed. Each original sequence is a segment of $\stream$. 
%We will avoid explicitly mentioning batched data, even if what we present in this paper is fine also in the case in which $\stream$ returns mini-batches.
%
%\parafango{Continuous Time} We will present our main results in a continuous time setting (time $\t$), that allows us to explicitly take care of the distance (in time) between consecutive samples from the stream, which might be uneven. However, we will also discuss discrete time, comparing the outcomes of the two settings. 
%The continuous-time $\stream$ yields data only at specific time instants $\t_1, \t_2, \ldots, \t_{\infty}$, evenly or not evenly spaced.

\parafango{Scheduling} We assume the most basic schedule of computations: the agent which implements our neural networks monitors the stream and, when a batch of data is provided, it starts processing it, entering a ``busy'' state, and leaving it when it is done. Data from $\stream$ is discarded when the agent is busy. At time $\t_j$, the agent is aware of the time $\Delta(t_j)$ that passed from the previously processed sample.

\parafango{Distributed Neural Computations}
The aforementioned notion of spatial locality can be structured in several different ways. %Given the DAG associated to a 
In a multi-layer feed-forward network, we can create groups of consecutive layers, and assume $\fxi$ returns the variation of the output values of the {\it last layer of each group}, thus the state $\h$ is not composed by all the neuron outputs of the net. Computations within each group happens instantaneously, and delays are only among groups. This weakly spatially local architecture corresponds to the one exploited to build pipeline parallelism in neural networks \cite{meloni}, when each group is processed on a different devices/GPUs. Interestingly, the notion of costate in HL is what allows the transfer of learning-related information  among groups, that in \cite{meloni} is stored into appositely introduced variables.

\parafango{Delay-free State Computation}
HL offers the basic tools also to distribute layer-wise computations over time in a fully customizable way, going beyond the ones showcased in Section~\ref{sec:good}, for example by interleaving delay-free computations with delayed weight updates.
%For example, it can handle models that have no delays in the propagation of information while updating the state, but delays in the way the model weights are updated. 
Consider the case of a feed-forward network with $\numlayers$ layers, where $\h$ is the output of {\it all} the neurons of the net and $\fxid$ is the function that computes such outputs.
The data is sequentially provided, $\u_0$, $\u_1$, $\ldots$, $\u_{\infty}$, and the network is expected to make predictions on each of them (samples are i.i.d. in this case). % and we can interleave each pairs of consecutive samples with additional steps required to propagate the signal through the network. 
We can introduce a time-variant state transition function $\fxi$, that now explicitly depends on time, without compromising the theoretical grounding of HL, since assuming this dependency is pretty natural in optimal control. 
Replacing time with indices $\k$, to simplify the notation, we have
\begin{align}
    {\fxi}(\u_{\lfloor \k / \numlayers \rfloor}, \h\argk, \xwxi\argk, \k) &= \indicator_{\mathrm{mod}(\kappa, \nu)} \odot  \tau^{-1}\left(-\h\argk + \fxid(\u_{\lfloor \k / \numlayers \rfloor}, \h\argk, \xwxi\argk) \right), \ \text{for } \k=0,1,\ldots,\infty,
    \label{eq:ffasd}
\end{align}
where time/step $\k$ is in the last argument, differently from what has been presented so far. We considered the model of Eq.~\ref{eq:euleraa}, being $\indicator_{\k}$ a binary indicator vector which is $1$ in the positions of the neurons of the $\k$-th layer and $\mathrm{mod}$ computes the remainder of the integer division. Each input sample is virtually repeated for $\numlayers$ steps, to wait for the signal to propagate through the network. At each step $\k$, only the portion of the state associated to a single layer is updated, due to the indicator function. Thus, after every $\k / \numlayers$ steps, $\h$ is composed of delay-free outputs of all the neurons of the network for a given input $\u_{\lfloor \k / \numlayers \rfloor}$. The model of Eq.~\ref{eq:ffasd} continuously updates weights over time, for each $\k$, using the forward-estimate of the costate. This means that while predictions are delay-free, there is a delay in the way weights are updated.

\section{Optimal Control Theory} \label{alessandro}
A control system consists of a pair $(\f,\controlset)$, where $\controlset \subset\realset^m$ is called the \emph{control set} and it is the set of admissible values of the control\footnote{Usually $\controlset$ is assumed to be closed even if in some cases, for instance when we want to choose $\controlset =\realset^m$ it is possible to trade the boundedness of $\controlset$ with the coercivity of the lagrangian (see~\cite{cannarsa2004semiconcave}
ch.~7, p. 214 for details).} and $\f\colon\realset^n\times\controlset\times[t_0,+\infty)\to\realset^n$ is a continuous function that is called the \emph{dynamics of the system}. Using the notation $(\t)$ instead of subscript $\t$ as in we did in the main paper, the \emph{state equation} associated to the 
system then is 
\begin{equation}\label{eq:state-eq}
\left\{\begin{aligned}
&\dot \x(t)= \f(\x(t),\boldgreek{\boldgreek{\alpha}}(t),t),\qquad \hbox{a.e. in $[t_0,+\infty)$}\\
&\x(t_0)=\bar{\mathrm{\bf x}}
\end{aligned}\right.
\end{equation}
where $t_0\in\realset$, $\bar{\mathrm{\bf x}}\in\realset^n$. The function $\boldgreek{\boldgreek{\alpha}}$, misurable, is what is usually called the \emph{control strategy} or \emph{control function} and we denote with 
$\x(\,\cdot\,; t_0,\bar{\mathrm{\bf x}},\boldgreek{\alpha})$ the unique solution\footnote{We are indeed assuming that for all $a\in \controlset$, $f(\cdot,a)$ is Lipschitz.} of Eq.~\eqref{eq:state-eq}. 

Then an \emph{optimal control} problem consists in choosing the control policy $\boldgreek{\alpha}$ in such a way that a functional, usually called \emph{cost} is minimized. In other words the objective of such class of
problems is to steer the dynamics of the system so that it performs ``well'' according to a certain criteria. In this paper we are interested in 
what is known as control problem of \emph{Bolza type}, in which, given a control system $(\f,\controlset)$, a terminal cost $g\in C(\realset^n)$ a time $N>0$ and a function 
$\ell(\cdot,\cdot,s)\in C(\realset^n\times\controlset\times[t,N];\realset)$ usually called \emph{lagrangian},
for all $(t,\tilde{\mathrm{\bf x}})\in[0,N]\times \realset^n$
the functional to be minimized is the total cost:
\begin{equation}\label{eq:cost2}
C_{t,\tilde{\mathrm{\bf x}}}(\boldgreek{\alpha}):=\int_t^N \hskip -3mm \ell(\x(s; t,\tilde{\mathrm{\bf x}},\boldgreek{\alpha}),\boldgreek{\alpha}(s),s)\, ds +g(\x(N; t,x,\boldgreek{\alpha})).
\end{equation}
Hence, more concisely the problem that optimal control (of Bolza type) is concerned with is 
\begin{equation}\label{eq:min-prob}
\inf C_{t,\tilde{\mathrm{\bf x}}}(\boldgreek{\alpha})
\end{equation}
over all possible control trajectories $\boldgreek{\alpha}\colon[t,N]\to\controlset$.

Here we briefly summarize some of the classical results that comes from the method of \emph{dynamic programming}
that are most relevant in this work and we refer to the excellent books~\cite{bardi1997optimal, cannarsa2004semiconcave} for additional details, 
and in general for a more precise and consistent definition of the theory.
The main idea behind dynamic programming is to embed the minimization problem~\eqref{eq:min-prob} into a larger class of such problems through the definition of the \emph{value function}:
\[
V(t,\tilde{\mathrm{\bf x}}):=\inf\{ C_{t,\tilde{\mathrm{\bf x}}}(\boldgreek{\alpha}) : \boldgreek{\alpha}\colon[t,T]\to\controlset\quad\hbox{is measurable}\}.
\]
In particular under suitable regularity conditions on the lagrangian, on the dynamics and on the terminal cost it is possible to show that the value function $V$ satisfies the following Hamilton-Jacobi
equation:
\begin{equation}\label{eq:H-J}
\begin{cases}
V_t(t,\tilde{\mathrm{\bf x}})+H(\tilde{\mathrm{\bf x}},\nabla V(t,\tilde{\mathrm{\bf x}}),t)=0 & (t,\tilde{\mathrm{\bf x}})\in(0,N)\times\realset^n\\
V(T,\tilde{\mathrm{\bf x}})=g(\tilde{\mathrm{\bf x}})& \tilde{\mathrm{\bf x}}\in\realset^n,
\end{cases}
\end{equation}
where the hamiltonian is defined as 
\begin{align}
    \nonumber H(\tilde{\mathrm{\bf x}},p,t)&=\min_{\mathrm{\bf a}\in \controlset} [p\cdot f(\tilde{\mathrm{\bf x}},\mathrm{\bf a},t)+L(\tilde{\mathrm{\bf x}},\mathrm{\bf a},t)] \\
    \nonumber &= -\max_{\mathrm{\bf a}\in \controlset} [-p\cdot f(\tilde{\mathrm{\bf x}},\mathrm{\bf a},t)-\ell(\tilde{\mathrm{\bf x}},\mathrm{\bf a},t)].
\end{align}
Once the value function is determined, the optimal control, i.e., the solution of the problem  can then be recovered 
by means of what is known as the \emph{synthesis procedure} (see~\cite{bardi1997optimal}). Basically once $V$
is known we can define for all $\tilde{\mathrm{\bf x}}\in\realset^n$ and for all $s\in[t,T]$ the \emph{optimal feedback map}
$S(\tilde{\mathrm{\bf x}},s)\in \argmin_{a\in\controlset} \nabla V(s,\tilde{\mathrm{\bf x}})\cdot f(\tilde{\mathrm{\bf x}},a,s)+\ell(\tilde{\mathrm{\bf x}},\mathrm{\bf a},s)$ and with $S$ compute the optimal trajectory 
$\x^*$
as a solution of $\dot \x(s)= \f(\x(s),S(\x(s),s),s)$ with initial condition $\x(t)=\bar{\mathrm{\bf x}}$. Then, finally, the optimal control 
$\boldgreek{\alpha}^*$ can be computed as $\boldgreek{\alpha}^*(s)=S(\x^*(s),s)$.

\parafango{Hamilton Equations} An alternative approach to the problem in Eq.~\eqref{eq:min-prob} relies on 
an alternative representation of the value function which is obtained through the method of characteristics
(see~\cite{courant2008methods})
and makes it possible to compute use a system of ODE instead of a PDE like Eq.~\eqref{eq:H-J} to construct the 
solution of the optimal control problem. This approach is also related to the  Pontryagin Maximum Principle~\cite{giaquinta2013calculus}.

Define the costate function as $\costate(s)=\nabla V(s,\x(s))$ and consider the following system of ODEs:
\begin{equation}\label{eq:Hamilton-eq}
\begin{cases}
\dot\x(s)=H_p(\x(s),\costate(s),s) & \hbox{for $s\in(t,N]$}\\
\dot\costate(s)=-H_x(\x(s),\costate(s),s)& \hbox{for $s\in(t,N]$}\\
\x(t)=\bar{\mathrm{\bf x}} & \text{\sc\small initial condition}\\
\costate(N)=\nabla g(\x(N)) & \text{\sc\small terminal condition},
\end{cases}
\end{equation}
where $H_x$ and $H_p$ are the partial derivatives of $H$ with respect to its first and second argument
respectively. Then if we are able to find the solution $(\x^*,\costate^*)$ of the system~\eqref{eq:Hamilton-eq},
then the optimal control can be directly computed as 
\[\boldgreek{\alpha}^*(s)\in \argmin_{\mathrm{\bf a}\in\controlset} \costate^*(s)\cdot f(\x^*(s),\mathrm{\bf a},s)+\ell(\x^*(s),\mathrm{\bf a},s).\]
This way, we converted the problem of solving Eq.~\ref{eq:min-prob} in the point-wise minimization problem of the equation above.

\section{Robust Hamiltonian and Forward Hamiltonian Equations}
\label{app:robust}
Considering the cost of Eq.~\ref{eq:cost}, the Hamiltonian function is the outcome of adding to the instantaneous cost the dot product between state and costate, evaluated at the minimum with respect to the control,
\begin{equation}
\begin{split}
    \nonumber \H(\h,\w, \pxi, \pw | \fw) &= \\
    \nonumber & \hskip -1cm = e^{\eta\t}C(\h, \w, \fw, t) + \pxi^T \fxi(\u, \xxi, \xwxi) + \pw^T(\beta\odot\fw) \\
    \nonumber & \hskip -1cm = e^{\dis\t}L\left(\fy(\u, \xxi, \xwy), \target\right)\phi + \frac{e^{\dis\t}}{2}\|\dotxw\|^2 + \pxi^T \fxi(\u, \xxi, \xwxi) + \pw^T(\beta\odot\fw)
   % & = e^{-\dis\t} L\left(\fy(\u, \xxi, \xwy), \target\right)\phiexp + \frac{1}{2}\|\dotxw\|^2_{\betaw} + \pxi^T \fxi(\u, \xxi, \xwxi) + \pw^T\dotxw,    
\end{split}
    \label{eq:hamreal}
\end{equation}
where we dropped the time index to keep the notation simple and where, from Eq.~\ref{eq:netsplitb}, $\dotxw = \betaw \odot \fw$. %We also considered $\phi$ defined as in Eq.~\ref{eq:phi}.
The definition of $\H$ remarks that, for each quadruple of arguments in Eq.~\ref{eq:hamreal}, we are in a stationary point w.r.t. to the control $\fw$. From now on, we completely avoid reporting the arguments of the involved functions, considering them to be the ones of Eq.~\ref{eq:hamreal}.
We get
\begin{equation}
    \frac{\partial \H}{\partial \fw} = e^{\dis\t}{\fw} + \pw \odot \betaw.
    \label{eq:opts}
\end{equation}
Setting Eq.~\ref{eq:opts} to zero (since it is known to be a stationary point), we get,
\begin{equation}
    \fw = - e^{-\dis\t}\beta \odot \pw.
    \label{eq:super}
\end{equation}
Once we use Eq.~\ref{eq:super} as value of the first argument of Eq.~\ref{eq:hamreal}, the Hamiltonian becomes,
\begin{equation}
    \begin{split}
    \H &= e^{\dis\t} L\phiexp + \frac{e^{\dis\t}}{2}\|e^{-\dis\t}\betaw \odot \pw\|^2 + \pxi^T \fxi - e^{-\dis\t}\pw^T (\betaw \odot \betaw \odot \pw) \\
    & = e^{\dis\t} L\phiexp + \frac{e^{-\dis\t}}{2}\|\betaw \odot \pw\|^2_2 + \pxi^T \fxi - e^{-\dis\t}\pw^T (\betaw \odot \betaw \odot \pw)
    \end{split}
    \label{eq:hamctrl}
\end{equation}
where we considered $\dotxw = \betaw \odot \fw = -e^{-\dis\t}\betaw \odot \betaw \odot \pw$ (merging its definition $\dotxw = \betaw \odot \fw$ with Eq.~\ref{eq:super}). Using the expression of $\H$ in Eq.~\ref{eq:hamctrl}, we can now compute the HEs,
\begin{align}
\dotxxi = \frac{\partial \H}{\partial \pxi} &= {\fxi} \tag{{\color{blue}$\mathcal{E}$$\star\tilde{1}$}} \\
\nonumber \dotxw = \frac{\partial \H}{\partial \pw} &= e^{-\dis\t}(\betaw \odot \betaw \odot \pw) - 2e^{-\dis\t} (\betaw \odot \betaw \odot \pw) \\
& = - e^{-\dis\t}(\betaw \odot \betaw \odot \pw) \tag{{\color{blue}$\mathcal{E}$$\star\tilde{2}$}} \\
\dotpxi = -s\frac{\partial \H}{\partial \h} &=  -se^{\dis\t} \phiexp \frac{\partial L}{\partial \h}  -s\pxi^T \frac{\partial \fxi}{\partial \h } \tag{{\color{red}$\mathcal{E}$$\star\tilde{3}$}}\\
\dotpw = -s\frac{\partial \H}{\partial \xw} &=  -se^{\dis\t} \phiexp \frac{\partial L}{\partial \w}  -s\pxi^T \frac{\partial \fxi}{\partial \w} \tag{{\color{red}$\mathcal{E}$$\star\tilde{4}$}}.
\end{align}
In order to avoid evaluating the exponential, we replace $\pxi$ with $e^{\dis\t}\hat{\pxi}$ and $\pw$ with $e^{\dis\t}\hat{\pw}$, which yield $\dotpxi = e^{\dis\t}\dot{\hat{\pxi}} + \dis e^{\dis\t}{\hat{\pxi}} $ and $\dotpw = e^{\dis\t}\dot{\hat{\pw}}+ \dis e^{\dis\t}{\hat{\pw}}$. Hence, the last two HEs can be written in function of the newly introduced variables $\hat{\pxi}$ and $\hat{\pw}$,
\begin{align}
\dotxxi  &= {\fxi}^T \tag{{\color{blue}$\mathcal{E}$$\star\tilde{1}$}} \\
\dotxw &= -\betaw \odot \betaw \odot \hat{\pw} \tag{{\color{blue}$\mathcal{E}$$\star\tilde{2}$}} \\
e^{\dis\t}\dot{\hat{\pxi}} + \dis e^{\dis\t}{\hat{\pxi}} &=  -se^{\dis\t} \phiexp \frac{\partial L}{\partial \h}  -se^{-\eta\t}\hat{\pxi}^T \frac{\partial \fxi}{\partial \h } \tag{{\color{red}$\mathcal{E}$$\star\tilde{3}$}}\\
e^{\dis\t}\dot{\hat{\pw}}+ \dis e^{\dis\t}{\hat{\pw}} &=  -se^{\dis\t} \phiexp \frac{\partial L}{\partial \w}  -se^{\dis\t}\hat{\pxi}^T \frac{\partial \fxi}{\partial \w} \tag{{\color{red}$\mathcal{E}$$\star\tilde{4}$}}.
\end{align}
which, also replacing $\betaw$ with $\hat\betaw^{1/2}$, can be simplified to
\begin{align}
\dotxxi  &= {\fxi} \tag{{\color{blue}$\mathcal{E}$$\star\tilde{1}$}} \\
\dotxw &= -\hat{\betaw} \odot \hat{\pw} \tag{{\color{blue}$\mathcal{E}$$\star\tilde{2}$}} \\
\dot{\hat{\pxi}} &=  -s \phiexp \frac{\partial L}{\partial \h}  -s\hat{\pxi}^T \frac{\partial \fxi}{\partial \h } - \dis {\hat{\pxi}} \tag{{\color{red}$\mathcal{E}$$\star\tilde{3}$}}\\
\dot{\hat{\pw}} &=  -s \phiexp \frac{\partial L}{\partial \w}  -s\hat{\pxi}^T \frac{\partial \fxi}{\partial \w} - \dis {\hat{\pw}} \tag{{\color{red}$\mathcal{E}$$\star\tilde{4}$}}.
\end{align}
Despite being defined as partial derivatives of the Hamiltonian of Eq.~\ref{eq:hamreal}, the first two HEs above can be computed without involving any differentiation. This suggest that we can avoid computing in $\H$ those terms that are responsible of yielding such first two HEs.
Considering the last two HEs, the first two terms in the right-hand side are the derivatives of
\begin{equation}
    L\phiexp + \hat{\pxi}^{T}\fxi
    \label{eq:hrestrictedx}
\end{equation}
with respect to $\h$ and $\w$, multiplied by $-s$. The offsets $\dis {\hat{\pxi}}$ and $\dis {\hat{\pw}}$ act like dissipation factors.
Eq.~\ref{eq:hrestrictedx} is the Robust Hamiltonian that was defined in Eq.~\ref{eq:hrestricted}. Comparing it with the ``real'' Hamiltonian of Eq.~\ref{eq:hamreal}, there is no exponentiation and the costate of the weights is not there at all, thus it is simpler/faster to compute. Finally, in order to get the HEs reported in the main paper, i.e., Eq.~\ref{eq:hesa}, Eq.~\ref{eq:hesb}, Eq.~\ref{eq:hesc}, Eq.~\ref{eq:hesd}, we just need to go back to the original notation based on $\betaw$, $\dotpxi$, $\pxi$, $\dotpw$, $\pw$ instead of $\hat\betaw$, $\dot{\hat\pxi}$, $\hat\pxi$, $\dot{\hat{\pw}}$. Notice that this is not a variable change, as we did before, but just a replacement of the notation.

\section{Out-of-the Box Tools vs. Hamiltonian Learning}
\label{oob}
{\it The code of HL and the script that runs all the comparisons is attached to the submission.}

\parafango{Converting Parameter Values} In our experimental comparison, we exploited out-of-the box  tools for gradient-based optimization in neural networks. The form of the update step in the case of gradient with momentum we considered
\begin{equation}
    \begin{split}
    \mathrm{\bf b}\argtnext &= \mu \mathrm{\bf b}\argt + (1 - \rho) \mathrm{\bf g}\argt \\
    \w\argtnext &= \w\argt + \lr (-\mathrm{\bf b}\argtnext)
    \end{split}
\end{equation}
where $\mathrm{\bf g}$ is the gradient of the loss with respect to $\w$ (with $\mathrm{\bf b}_{\tau} = \mathrm{\bf g}_0$), while $\mu$, $\rho$, $\lr$ are the momentum term, the dampening factor, and the learning rate, respectively, coherently with the Pytorch implementation of the SGD optimizer\footnote{\url{https://pytorch.org/docs/stable/generated/torch.optim.SGD.html}}, that is what we used in our comparisons (time indices are adjusted following the notation of this paper).

After having selected any $\eulerstep > 0$, we can map the optimizer parameters onto the parameters of Hamiltonian Learning as in the following table:

\begin{minipage}{1.0\columnwidth}
\begin{center}
\begin{tabular}{c|c}
    \textsc{\small Pytorch Optim.} & \textsc{\small Hamiltonian Learning}  \\
    \hline
    $\lr$ & $\betaw \eulerstep$ \\
    $\mu$ & $(1-\eulerstep \dis)$ \\
    $\rho$ & $1 - \eulerstep \phiexp$ 
\end{tabular}
\end{center}
\end{minipage}
from which we have $\beta = \lr / \eulerstep$, $\dis=(1-\mu) / \eulerstep$ and $\phiexp = (1 - \rho) / \eulerstep$. In Section~\ref{sec:grad} we compared the following cases:

\begin{minipage}{1.0\columnwidth}
\begin{center}
\hskip -0.7cm
\begin{tabular}{c|cccc}
    \textsc{\small Pytorch Optim.} & \textsc{\small GD-a} & \textsc{\small GD-b} & \textsc{\small Mom-a} & \textsc{\small Mom-b}  \\
    \hline
    $\lr$ & \small $0.01$ & \small $0.001$ &\small $0.01$ & \small $0.01$ \\
    $\mu$ & \small $0$ & \small $0$ &\small $0.05$ & \small $0.1$ \\
    $\rho$ & \small $0$ & \small $0$ &\small $0.6$ & \small $0.5$
\end{tabular}
\end{center}
\end{minipage}
while we set $\eulerstep$ to $1$, $0.5$, $1$, $0.5$ in the four cases of the table above, respectively.

\parafango{Data \& Setup} In this study, the batch size was consistently set to 1. We exploited a subsample of  Iris\footnote{Iris dataset is licensed under a Creative Commons Attribution 4.0 International (CC BY 4.0) license. \url{https://archive.ics.uci.edu/dataset/53/iris}} \cite{misc_iris_53}, MNIST \footnote{\url{http://yann.lecun.com/exdb/mnist/}} and the IMDb datasets \cite{imdb}. The Iris dataset included all 150 samples across its 4 classes, with training conducted over 40 epochs. From the MNIST dataset, 100 test set samples were used, with 10 samples per class across 10 classes, also trained for 40 epochs. For the IMDb dataset, 100 training set samples were selected, split evenly between its 2 classes, and trained for 80 epochs due to the more challenging convergence. Experiments were executed on a Linux server with Pytorch 2.3.0, equipped with a 10th-Gen Intel CPU and 64GB of RAM (experiments take minutes to run, log, and plot results).

% - MNIST http://yann.lecun.com/exdb/mnist/
% - Iris This dataset is licensed under a Creative Commons Attribution 4.0 International (CC BY 4.0) license. https://archive.ics.uci.edu/dataset/53/iris
% - IMDb https://ai.stanford.edu/~amaas/data/sentiment/ (\cite{imdb}
% Experiments executed on a Linux server with Pytorch 2.3.0, equipped with a 10th-Gen Intel CPU and 64GB of RAM (experimens take minutes to run, log, and plot results).

\parafango{Models \& Metrics} We evaluated six different model architectures: a single-layer neural network (Linear), a multi-layer perceptron with one hidden layer of 30 neurons using the tanh activation function (MLP), a small ResNet comprising 4 residual blocks (ResNet), a simple Vision Transformer with 1 encoder layer (30 hidden neurons, patch size 7x7), 1 layer normalization, 1 feedforward network (30 to 30 neurons), and 1 head (ViT), a recurrent neural network using pre-trained GloVe embeddings (50-dimensional), with a hidden state of 30 neurons and tanh activation, followed by a linear layer for classification (RNN), and a similar architecture to the RNN but with an LSTM unit instead of the RNN (LSTM).  We report each model performance using the loss function values (cross-entropy) and accuracy metrics, during the model training phase.

\begin{figure*}
    \centering
% First Row
\includegraphics[width=0.25\linewidth,trim={2cm 2cm 2cm 2cm},clip]{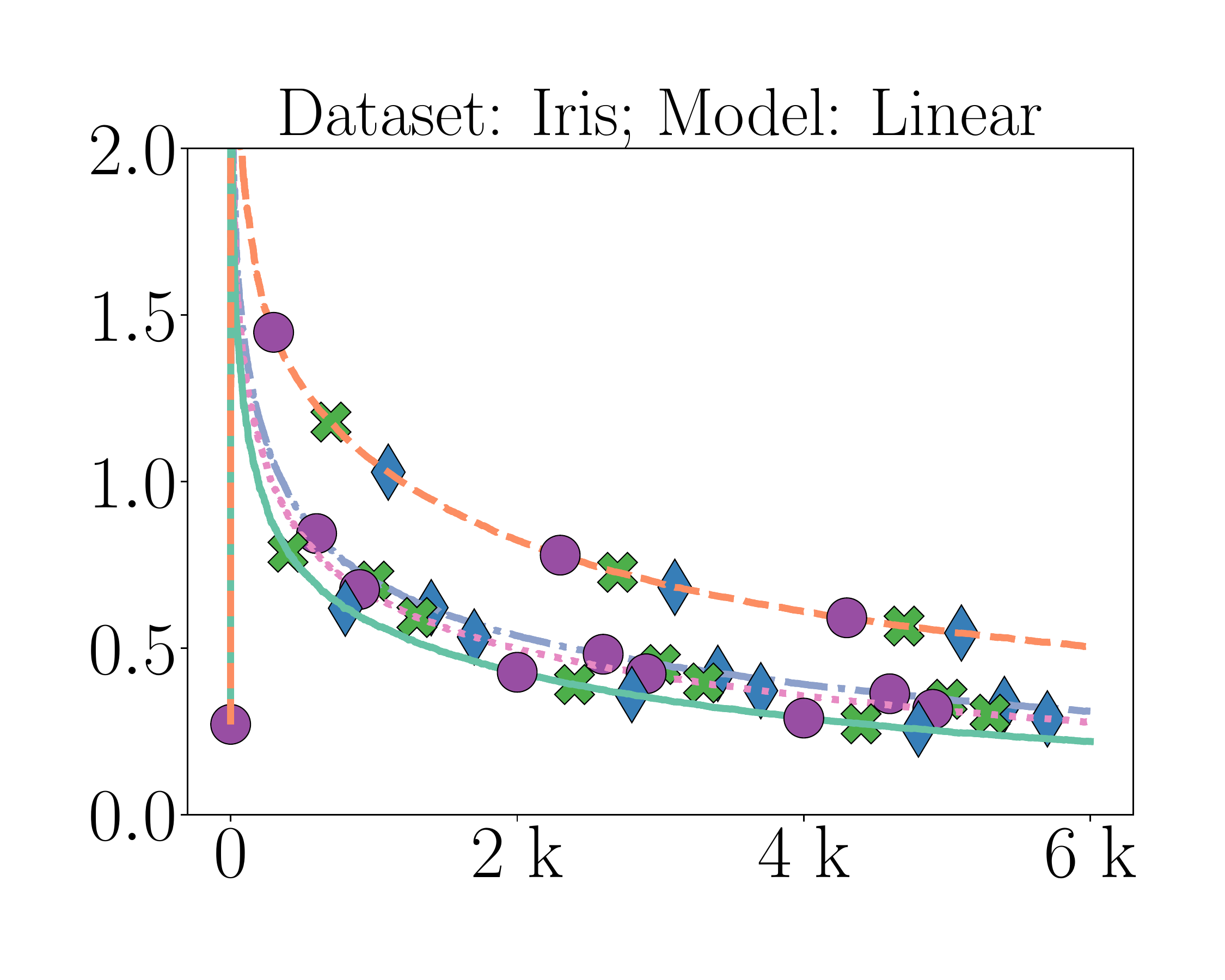} 
\includegraphics[width=0.25\linewidth,trim={2cm 2cm 2cm 2cm},clip]{figs/MNIST_ResNet_loss.pdf} 
\includegraphics[width=0.25\linewidth,trim={2cm 2cm 2cm 2cm},clip]{figs/IMDb_RNN_loss.pdf} \\
% Second Row
\includegraphics[width=0.25\linewidth,trim={2cm 2cm 2cm 2cm},clip]{figs/Iris_MLP_loss.pdf} 
\includegraphics[width=0.25\linewidth,trim={2cm 2cm 2cm 2cm},clip]{figs/MNIST_ViT_loss.pdf} 
\includegraphics[width=0.25\linewidth,trim={2cm 2cm 2cm 2cm},clip]{figs/IMDb_LSTM_loss.pdf} \\
    \includegraphics[width=0.6\linewidth,trim={0cm 15cm 0 0},clip]{figs/legend.pdf}
    \caption{Experimental comparisons, Loss Values. We report the outcome of comparing gradient-based learning (with and without momentum, denoted with \textsc{GD} and \textsc{Mom}, respectively) using popular out-of-the-box tools (we tested two different configurations, denoted with the suffix "\textsc{-a}" and "\textsc{-b}", respectively, having different learning rates, momentum terms, damping factors) and Hamiltonian Learning (\textsc{HL}, setting $\tau$, $\beta$, $\dis$, $\phi$ to values that we theoretically show to be coherent with the parameters of out-of-the-box tools-- see Appendix~\ref{oob}). When considering \textsc{HL}, we can implement the selected model by solely considering the output function (output), the state function (state), or we can split it putting a portion into the state and a portion into the output (state/output). The plot shows the perfect alignment in terms of cost function values  during the models' training phase ($x$-axis, training steps).}
    \label{fig:exp2}
\end{figure*}

\begin{figure*}
    \centering
    % Third Row
\includegraphics[width=0.25\linewidth,trim={2cm 2cm 2cm 2cm},clip]{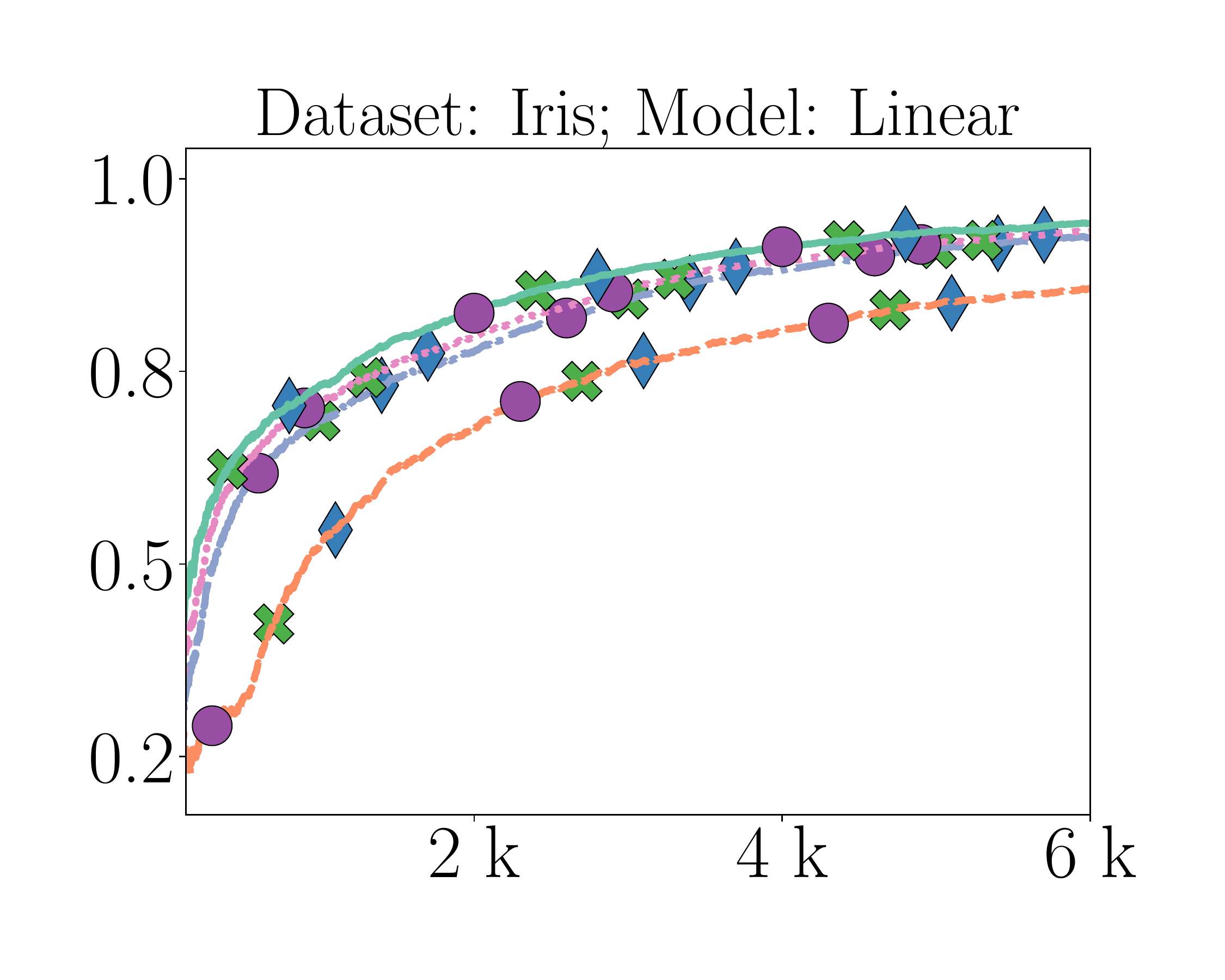} 
\includegraphics[width=0.25\linewidth,trim={2cm 2cm 2cm 2cm},clip]{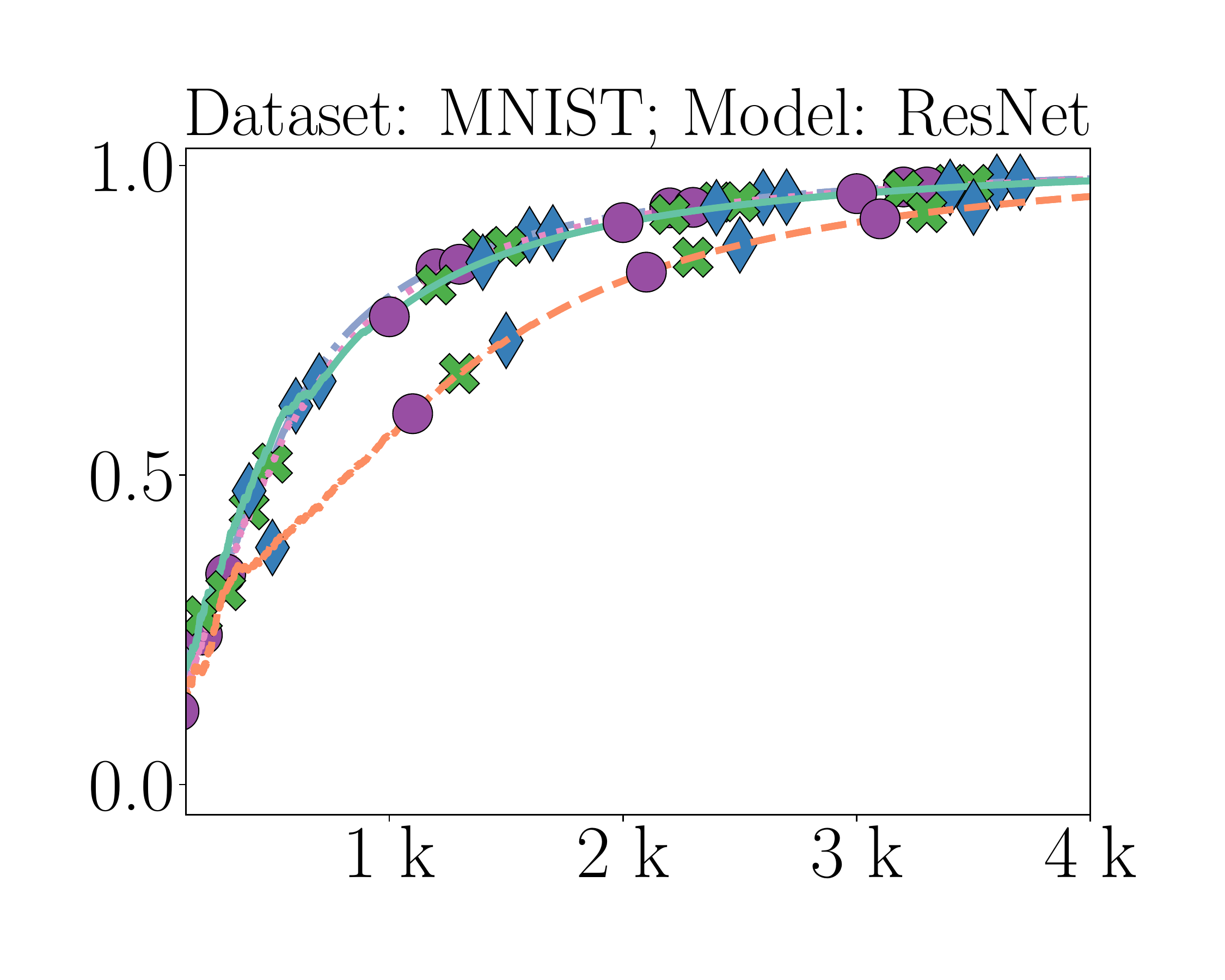} 
\includegraphics[width=0.25\linewidth,trim={2cm 2cm 2cm 2cm},clip]{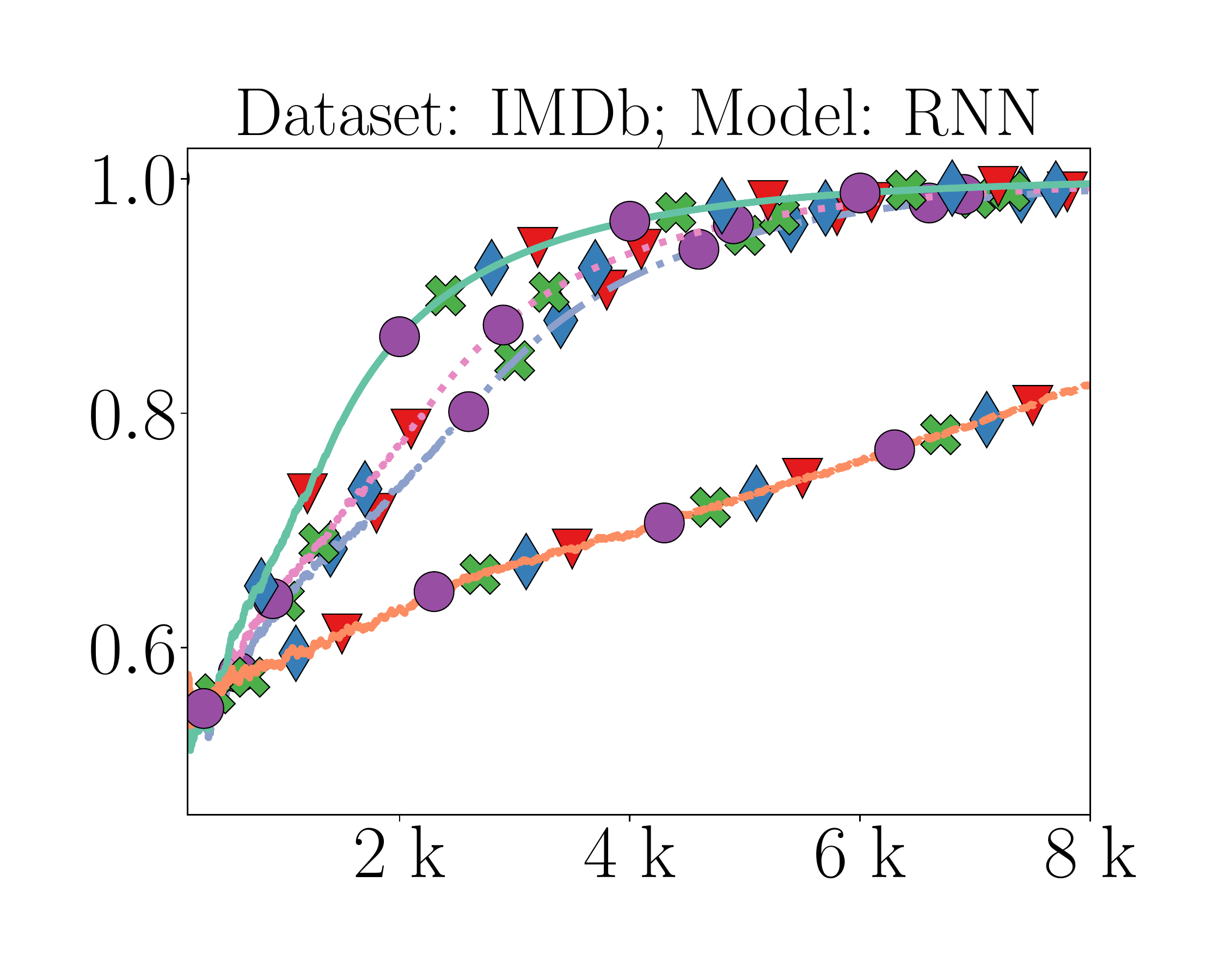} \\
% Fourth Row
\includegraphics[width=0.25\linewidth,trim={2cm 2cm 2cm 2cm},clip]{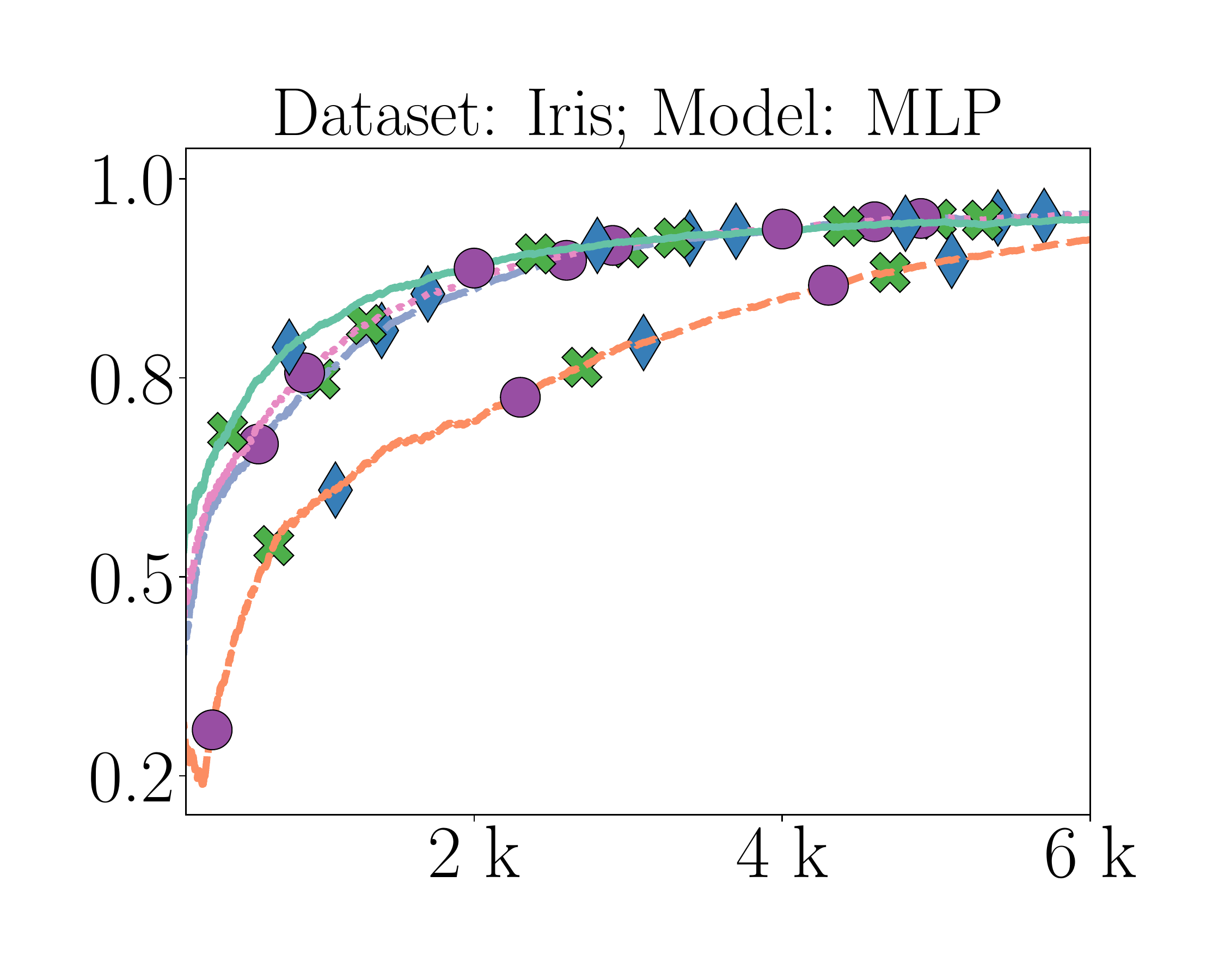} 
\includegraphics[width=0.25\linewidth,trim={2cm 2cm 2cm 2cm},clip]{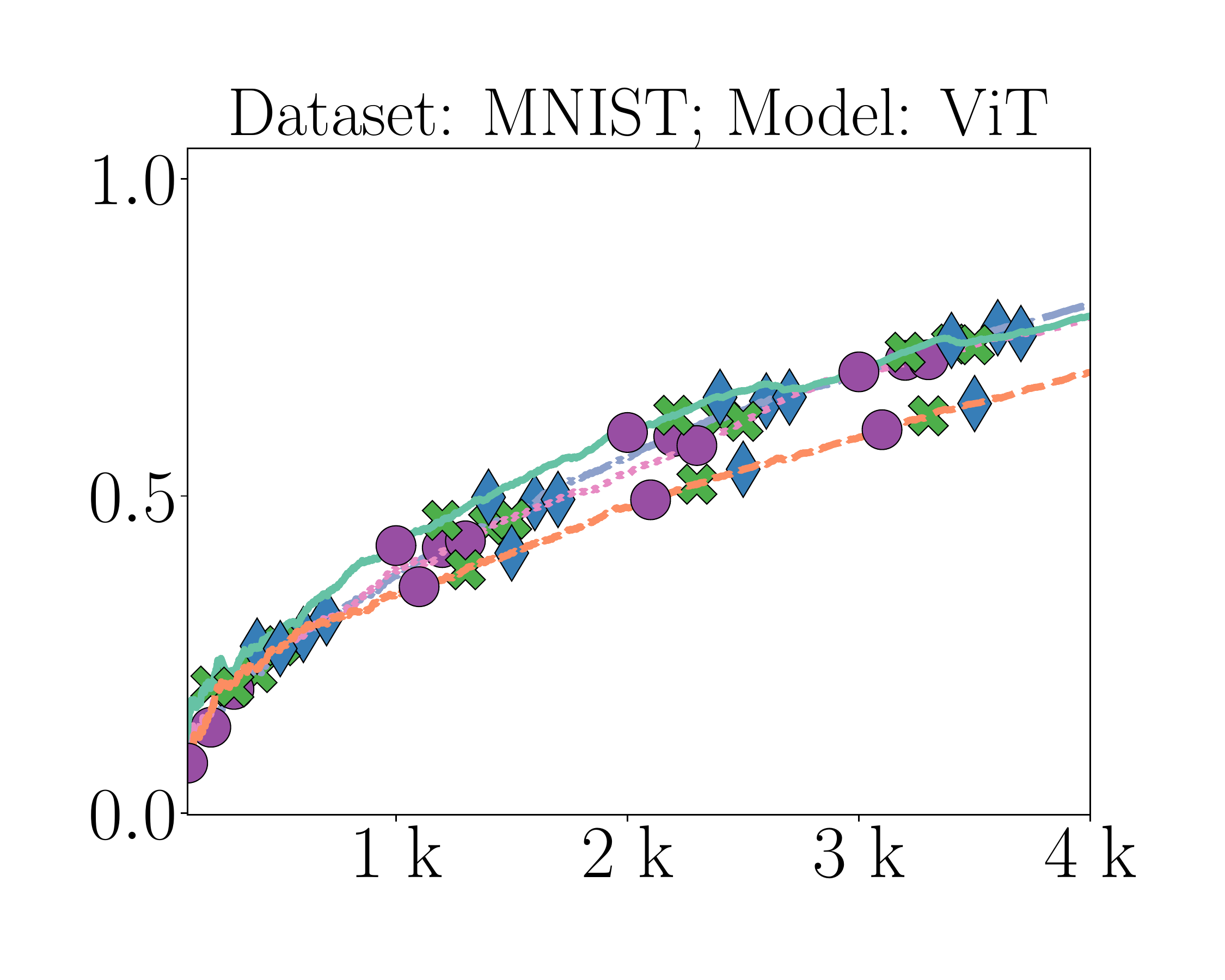} 
\includegraphics[width=0.25\linewidth,trim={2cm 2cm 2cm 2cm},clip]{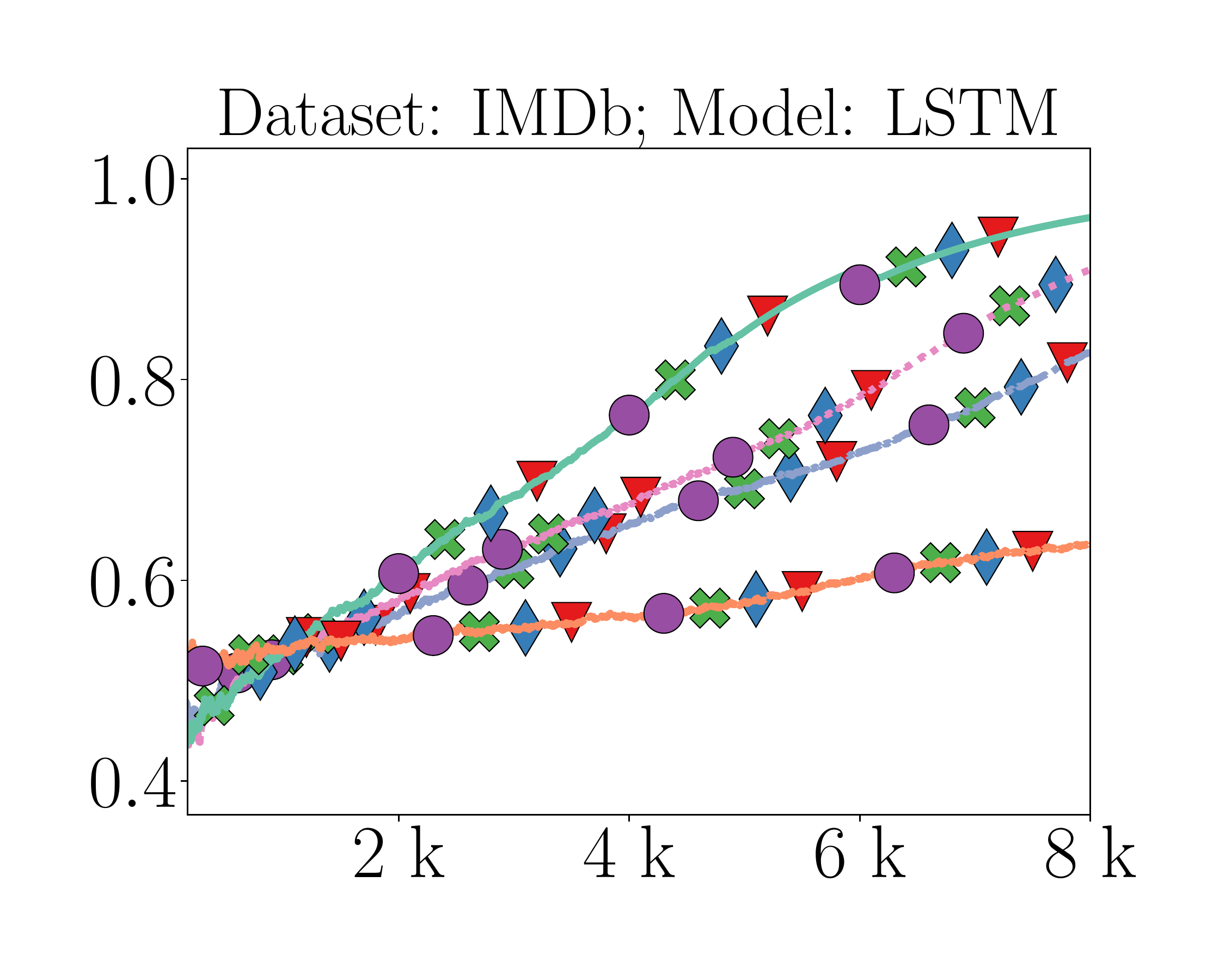} \\
    \includegraphics[width=0.6\linewidth,trim={0cm 15cm 0 0},clip]{figs/legend.pdf}
    \caption{Experimental comparisons, Accuracy. Same setting of Figure \ref{fig:exp2}. The plot shows the perfect alignment in terms of accuracy values  during the models' training phase ($x$-axis, training steps).}
    \label{fig:exp3}
\end{figure*}

\parafango{Results} Figures \ref{fig:exp2}-\ref{fig:exp3} report the attained results in terms of loss values and accuracy, respectively,  comparing the proposed HL method with several out-of-the-box gradient descent methods.
The first and second rows illustrate the loss values for the different tested models, across the datasets (columns). For the Iris dataset (first column), both the Linear model and the MLP show that the proposed HL method (represented by green crosses and blue diamonds) aligns perfectly with the loss value of the out-of-the-box gradient descent (GD) methods (purple circles) in all the tested learning configurations (\textsc{GD-a/b} and \textsc{Mom-a/b} -- various dashed lines). The same trend is observed in the MNIST dataset (second column) for the ResNet and ViT models, where HL maintains a loss value comparable to other GD methods. On the IMDb dataset (third column), both the RNN and LSTM models using HL also exhibit identical loss convergence dynamics to the ones of established GD techniques.
Figure \ref{fig:exp3} focuses on accuracy. In the Iris dataset, the linear model and MLP achieve accuracy levels with HL that are consistent with those of the other methods. This consistency is further demonstrated in the MNIST dataset for the ResNet and ViT models, where HL's accuracy exactly recovers the one of other gradient descent approaches. For the IMDb dataset, both the RNN and LSTM models show that HL achieves accuracy identical to the traditional methods.

\parafango{Quantitative Numerical Differences} The starting point of the optimization is the same for all the compared cases, of course. We measured the numerical difference in weights values at the final training step, comparing the values attained by proposed HL method and the out-of-the-box gradient descent methods. In details, we compared the mean, min, max absolute differences of weights. It is numerically zero in most cases or within the order of round-off error. The \textsc{Mom-a} learning configuration is the only one which presented barely measurable differences, that are in the order of $\approx 1.0e^{-10}$ for the mean and absolute weight value difference, for all the considered models and settings.
This indicates that HL produces the same model parameters, further supporting how it can generalize traditional optimization methods.
%Overall, the results indicate that the proposed HL framework performs equivalently to out-of-the-box gradient descent methods, both in minimizing loss and maximizing accuracy across various models and datasets. This demonstrates the robustness and reliability of the HL approach in different experimental settings.

\section{Feed-forward Networks and State Net}
\label{ffandstate}
When implementing a feed-forward network using the state network, we rely on the instantaneous propagation of Eq.~\ref{eq:euleraa}. We get
\begin{equation}
    \begin{split}
    \h\argtnext &= \h\argt + \eulerstep \cdot \eulerstep^{-1}\left(-\xxi\argt + \fxid(\u\argt, \xxi\argt, \xwxi\argt) \right)\\
    &= \fxid(\u\argt, \h\argt, \xwxi\argt)
    \end{split}
    \label{eq:figliodi1}.
\end{equation}
In order to avoid dependencies from the past, we clear the state, $\h\argt = {\mathrm{\bf 0}}$ and,
since $\dotxxi\argt = \eulerstep^{-1}(-\xxi\argt + \fxid(\u\argt, \xxi\argt, \xwxi\argt))$, we get $\dotxxi\argt = \eulerstep^{-1}\fxid(\u\argt, \xxi\argt, \xwxi\argt)$. Comparing this last equation with Eq.~\ref{eq:figliodi1}, we obtain
\begin{equation}
    \begin{split}
    \h\argtnext &=  \fxid(\u\argt, \xxi\argt, \xwxi\argt) = \eulerstep\dotxxi\argt
    \end{split}
    \label{eq:figliodi11}.
\end{equation}
which is the same as Eq.~\ref{eq:figliotop}. Finally, Eq.~\ref{eq:figliobottom} is trivially obtained from Eq.~\ref{eq:hesc}, setting $\pxi\argt = \mathrm{\bf 0}$, to avoid propagating past information.

\section{Learning in Recurrent Networks}
\label{bptt}
Learning in Recurrent Neural Networks (RNNs) is usually instantiated with the goal of minimizing a loss function defined on a window of sequential samples whose extremes are $a$ and $b$, with $b\geq a$, such as $\sum_{\k=a}^{b} L(\h\argk, \target\argk)$, where $\h\argk$ is the hidden state at step $\k$. Weights are constant while processing the data of the window. Let us consider the case of a dataset of sequences, sampled in a stochastic manner, and let us assume that at a certain stage of the optimization the values of the weights of the RNN are indicated with $\w$, while $a=1$ and $b$ is the sequence length. 
Once gradients $\partial \sum_{\k=a}^{b} L(\h\argk, \target\argk) / \partial \w$ are computed, weights are updated following the direction of the negative gradient.

Even if we avoided explicitly showing the dependence of $L$ on $\w$, to keep the notation simple, the loss function at step $\k$, i.e., $L(\h\argk, \target\argk)$, depends on $\w$ both due to its direct involvement into the computation of the current hidden state $\h\argk$ and also through all the previous states, up to $\h_{a}$.
When applying the chain rule, following the largely known BackPropagation Through Time (BPTT) \cite{pascanu2013difficulty}, we get,
\begin{equation}
\begin{split}
    \frac{\partial \sum_{\k=a}^{b} L(\h\argk, \target\argk)}{\partial \w} &= \sum_{\k=a}^{b} 
 \frac{\partial{L(\h\argk, \target\argk)}}{\partial{\w}} \\
 & = \sum_{\k=a}^{b} \sum_{q=a}^{\k} 
 \frac{\partial{L(\h\argk, \target\argk)}}{\partial\h\argk} \frac{\partial\h\argk}{\partial{\h_{q}}} \frac{\partial\h_{q}}{\partial{\w}} \\ &=\sum_{q=a}^{b} \left( \sum_{\kappa=q}^{b} 
 \frac{\partial{L(\h\argk, \target\argk)}}{\partial\h\argk} \frac{\partial\h\argk}{\partial{\h_{q}}} \right)\frac{\partial\h_{q}}{\partial{\w}},
\end{split}
 \label{eq:uffa}
\end{equation}
where in the last derivation we swapped the order of the two sums in $\k$ and $q$, adjusting the extremes of the summations. This can be easily done while noticing that the two summations in $\k$ and $q$ are represent the sum of the elements on a triangular matrix (lower triangular, if $\k$ is the row index), thus we can sum on rows or columns first, interchangeably.
We grouped some terms in brackets since the ${\partial\h_{q}}/{\partial{\w}}$ does not depend on index $\k$. Such terms can be iteratively evaluated going backward in time, exploiting the following relation,
\begin{equation}
    \frac{\partial \h_b}{\partial \h_q } = \prod_{\k=q+1}^{b} \frac{\partial \h_\k}{\partial \h_{\k-1}}
    \label{pascanumerda}
\end{equation}
with $q < b$, that can be used to compute the gradient w.r.t. $\w$ in an iterative manner, going over the input sequence in a backward manner. In fact, from Eq.~\ref{pascanumerda} we get
\begin{equation}
    \frac{\partial \h_b}{\partial \h_{q} } = \frac{\partial \h_b}{\partial \h_{q+1} }\frac{\partial \h_{q+1}}{\partial \h_{q} },
\end{equation}
which, starting from $q = b-1$, can be efficiently evaluated for $q=b-1,b-1,\ldots,a$.
We introduce additional variables to progressively accumulate gradients, i.e., $\tilde{\h}$ and $\tilde{\w}$, with $\tilde{\h}_{b} = \mathrm{\bf 0}$ and $\tilde{\w}_{b} = \mathrm{\bf 0}$. For $q=b,b-1,\ldots,a$ (here we have to consider also the gradient of the loss, that is why $q$ starts from $b$), we have
\begin{align}
    \tilde{\h}_{q-1} &= \frac{\partial L(\h_{q}, \target\argk)}{\partial \h_{q}} + \tilde{\h}_{q}^T \frac{\partial \h_{q+1}}{\partial \h_{q}}  \label{eq:bptt1}\\
    \tilde{\w}_{q-1} &= \tilde{\w}_{q} + \tilde{\h}_{q-1}^T \frac{\partial\h_{q}}{\partial{\w}},
    \label{eq:bptt2}
\end{align}
where ${\partial\h_{q}}/{\partial{\w}}$ can be immediately evaluated without involving data of the other time instants. Eq.~\ref{eq:bptt2} implements the (backward) summation on index $q$ of Eq.~\ref{eq:uffa} (bottom), where the term in brackets of Eq.~\ref{eq:uffa} (bottom) is $\tilde{\h}_{q}$. The summation in such a term in brackets is what is evaluated by Eq.~\ref{eq:bptt1}.
Finally,
\begin{equation}
\frac{\partial \sum_{\k=a}^{b} L(\h\argk, \target\argk)}{\partial \w} = \tilde\w_{a}.
\end{equation}

We can compare the BPTT equations Eq.~\ref{eq:bptt1} and Eq.~\ref{eq:bptt2} with the ones from Hamiltonian Learning, Eq.~\ref{eq:hescbptt} and Eq.~\ref{eq:hesdbptt}. 
Before going into details of the comparison, we mention that, to get Eq.~\ref{eq:hescbpttx}, we exploited
\begin{equation}
\begin{split}
    \frac{\partial \dot\h\argtnext}{\partial \h\argtnext} &= \eulerstep^{-1}\frac{\partial (-\h\argtnext+\fxid(\u\argtnext, \h\argtnext, \xwxi\argtnext))}{\partial \h\argtnext} \\
    &= \eulerstep^{-1} \left(-I + \frac{\partial \h_{\t+2\eulerstep}}{\partial \h\argtnext}\right),
\end{split}
\end{equation}
and we considered $s = 1$. Similarly, to get Eq.~\ref{eq:hesdbpttx} we exploited
\begin{equation}
    \frac{\partial \dot\h\argt}{\partial \xwxi\argt} = \eulerstep^{-1}\frac{\partial (-\h\argt+\fxid(\u\argt, \h\argt, \xwxi\argt))}{\partial \xwxi\argt} = \eulerstep^{-1} \frac{\partial \h_{\t+\eulerstep}}{\partial \xwxi\argt},
\end{equation}
since $\h\argt$ does not depend on $\xwxi\argt$ (again, $s=-1$).
%in order to  get Eq.~\ref{eq:hescbptt}, we exploited 
%$\partial\dot\h\argpsit/\partial\h\argpsit = \eulerstep^{-1}(-\partial\h\argpsit/\partial\h\argpsit+\partial\h\argpsitplusnext/\partial\h\argpsit) = \eulerstep^{-1}(-\partial\h\argpsit/\partial\h\argpsit+\partial\h\argpsitprev/\partial\h\argpsit) = \eulerstep^{-1}(-I + \partial\h\argpsitprev/\partial\h\argpsit)$, since $\psi\argtprev = \psi\argt + \eulerstep$ and $\psi\argtnext = \psi\argt - \eulerstep$.
%%
%Similarly, in Eq.~\ref{eq:hesdbptt}, after having evaluated $\pxi$ and the derivative $\dot\h$ in $\tnext$, coherently with what we did in the case of feed-forward networks, we exploited 
%$\partial\dot\h\argpsitnext/\partial\xwxi\argt = \eulerstep^{-1}(-\partial\h\argpsitnext/\partial\xwxi\argt+\partial\h\argtnextplusnext/\partial\xwxi\argt) = \eulerstep^{-1}(-\partial\h\argpsitnext/\partial\xwxi\argt+\partial\h\argpsit/\partial\xwxi\argt) = \eulerstep^{-1}\partial\h\argpsit/\partial\xwxi\argt$, since only the rightmost term in the summation directly depends on $\xwxi$ for the way we defined Eq.~\ref{eq:euleraa}.
Going back to the comparison between the BPTT equations Eq.~\ref{eq:bptt1} and Eq.~\ref{eq:bptt2} and the ones from Hamiltonian Learning, Eq.~\ref{eq:hescbptt} and Eq.~\ref{eq:hesdbptt}, 
we notice that the gradient accumulators $\tilde{\h}$ and $\tilde{\w}$ play the same role of the costates $\pxi$ and $\pwxi$, respectively. 
BPTT proceeds backward, while Hamiltonian Learning goes forward, processing the  sequence in reverse order. Thus, $\pxi\argtnext$ and $\pxi\argt$ correspond to $\tilde\h_{q-1}$ and $\tilde\h_{q}$, respectively (same comment for the case of $\pwxi$ and $\tilde\w$). 
%$\h\argpsit$ corresponds to $\tilde\h_{q-1}$ and $\tilde{\h}_q$ (same comment for the case of $\pwxi$ and $\tilde\w$). 
In Hamiltonian Learning, we need to recover the states stored when processing the original sequence, thus all the $\h$ terms involve map $\psi$ applied to time $\t$.
%In the case of $\h_{q+1}$, replacing $q+1$ with $\t - \eulerstep$ and evaluating map $\psi$, we get $\h_{\psi_{t-\eulerstep}}$. 
Target $\target\argk$ is associated to $\target_{\psi_{\t}-\eulerstep}$, showing that the time index of $\target$ is precedent with respect to the one of $\h$ in the loss function. This is coherent with what we did when implementing feed-forward nets with the state network, Eq.~\ref{eq:figliobottom}, where we have the same offset between the times of $\h$ and $\target$.
Setting $\phiexp\argt = \eulerstep^{-1}$, and no dissipation ($\dis = 0$), the correspondence between the two pairs of equations is perfect (i.e. the pairs of BPTT equations and the pairs of the costate-related Hamiltonian Learning equations).

\def\vlineA{\vskip-4mm$\left.\vphantom{\frac{1}{\sqrt{2}}}\right\vert$}
\def\vlineB{\vskip-3mm$\left.\vphantom{\frac{1}{\sqrt{2}}}\right\vert$}

\end{document}